\documentclass[lettersize,journal]{IEEEtran}
\usepackage{amsmath,amsfonts}
\usepackage{algorithmic}
\usepackage{algorithm}
\usepackage{array}
\usepackage[caption=false,font=normalsize,labelfont=sf,textfont=sf]{subfig}
\usepackage{textcomp}
\usepackage{stfloats}
\usepackage{url}
\usepackage{verbatim}
\usepackage{graphicx}
\usepackage{cite}
\usepackage[pagebackref,breaklinks,colorlinks,allcolors=blue]{hyperref}

\usepackage{xspace} % 加载 xspace 宏包
\usepackage{caption}

\usepackage{multirow}
\usepackage{booktabs}
\usepackage{pifont}
\usepackage[table,xcdraw]{xcolor}
\usepackage{amsmath}
\usepackage{amssymb}

\def\sc{\textcolor{magenta}}

\newcommand{\eg}{e.g.\xspace}
\newcommand{\ie}{i.e.\xspace}
\newcommand{\etc}{etc.\xspace}

\newcommand{\vs}{vs.\xspace}

\begin{document}

\title{ZOTTA: Test-Time Adaptation with Gradient-Free \\Zeroth-Order Optimization}

\author{Ronghao Zhang, Shuaicheng Niu, Qi Deng, Yanjie Dong, Jian Chen, Runhao Zeng
\thanks{Ronghao Zhang, Qi Deng and Jian Chen are with South China University of Technology, Guangzhou, 510000, China. 
Shuaicheng Niu is with Nanyang Technological University, 639798, Singapore. 
Yanjie Dong is with Xi'an Jiaotong University, Xi'an, 710049, China. 
Runhao Zeng is with Shenzhen MSU-BIT University, Shenzhen, 518172, China.
E-mail: zhangronghao16@gmail.com; shuaicheng.niu@ntu.edu.sg; dengqi.kei@gmail.com; ydong@xjtu.edu.cn; ellachen@scut.edu.cn; runhaozeng.cs@gmail.com.}}

% \markboth{IEEE Transactions on Circuits and Systems for Video Technology}%


\maketitle

\begin{abstract}
Test-time adaptation (TTA) aims to improve model robustness under distribution shifts by adapting to unlabeled test data, but most existing methods rely on backpropagation (BP), which is computationally costly and incompatible with non-differentiable models such as quantized models—limiting practical deployment on numerous edge devices. Recent BP-free approaches alleviate overhead but remain either architecture-specific or limited in optimization capacity to handle high-dimensional models. We propose ZOTTA, a fully BP-free TTA framework that performs efficient adaptation using only forward passes via Zeroth-Order Optimization (ZOO). While ZOO is theoretically appealing, naive application leads to slow convergence under high-dimensional parameter spaces and unstable optimization due to the lack of labels. ZOTTA overcomes these challenges through 1) Distribution-Robust Layer Selection, which automatically identifies and freezes layers that already extract distribution-invariant features, updating only domain-sensitive layers to reduce the optimization dimensionality and accelerate convergence; 2) Spatial Feature Aggregation Alignment, which stabilizes ZOO by aligning globally aggregated spatial features between source and target to reduce gradient variance. Together, these components enable architecture-agnostic and stable BP-free adaptation. Extensive experiments on ImageNet-C/R/Sketch/A show that ZOTTA outperforms or matches BP-based methods, e.g., it reduces memory usage by 84\% and improves accuracy by 3.9\% over SAR on ImageNet-C. Code: \href{https://github.com/Zhang-Ronghao/zotta}{https://github.com/Zhang-Ronghao/zotta}.
\end{abstract}


\section{Introduction}
\label{sec:intro}

Test-time adaptation (TTA) aims to improve model generalization under distribution shifts by adapting to unlabeled target-domain data during inference. This is critical for real-world deployment, where test distributions often change dynamically. Most existing TTA methods~\cite{TTT,TENT,MEMO,SAR,EATA} rely on backpropagation (BP) to update model parameters through auxiliary objectives such as self-supervision or entropy minimization. While effective, BP-based methods incur high computational and memory costs and cannot be applied to non-differentiable models, such as quantized models with vanishing gradients~\cite{DBLP:conf/iclr/LouizosRBGW19}, limiting their practicality in resource-constrained settings such as edge devices.

To bypass these limitations, BP-free TTA has gained interest. Learning-free approaches such as BN-Adapt~\cite{norm} and T3A~\cite{T3A} adjust feature statistics or prototypes for quick adaptation but offer limited performance gains. Learning-based methods, e.g., FOA~\cite{foa}, optimize lightweight prompts for Vision Transformers (ViTs) via evolutionary search, achieving better results but remaining architecture-specific. Overall, current BP-free methods still face two fundamental limitations (summarized in Table~\ref{table:bp_based_free}): \textit{1) Poor architectural generality:} Methods like BN-Adapt are restricted to CNNs with BatchNorm, while FOA relies on ViT architectures; \textit{2) Low optimization capacity:} Current designs can only adjust low-dimensional parameters, limiting scalability to complex, high-dimensional models.

\begin{table*}[t]
\centering
% \vspace{-0.15pt}
\renewcommand{\arraystretch}{1.0} % 改变行间距
\setlength{\tabcolsep}{10pt}% 改变列间距

\caption{Comparison of our method with previous backpropagation (BP)-based and BP-free test-time adaptation methods. Runtime accuracy and memory are the average results of the full-precision ViT-Base and ResNet50-GN models on ImageNet-C (corruption level 5) with batch size 64. FOA results are reported under the same number of forward passes as ours.}
% \vspace{-3pt}
% \footnotesize
\small

\resizebox{\textwidth}{!}{
\begin{tabular}{l|ccc|c}
\multirow{2}{*}{Method} & BP-Based & BP-Free \& Learning-Free & BP-Free \& Learning-Based  & \multirow{2}{*}{Ours} \\
                        & (\eg, SAR~\cite{SAR})    & (\eg, T3A~\cite{T3A})    & (\eg, FOA~\cite{foa}) &                       \\ \midrule
% Method & BP-Based (\eg, SAR~\cite{SAR})       &  BP-Free \& Learning-Free  (\eg, T3A~\cite{T3A})     & FOA~\cite{foa}      & Ours            \\    \midrule
Rely on BP?                                                               & Yes              & No                   & No        & No               \\
% \rowcolor[HTML]{EFEFEF} 
Learn from an objective?                                                         & Yes              & No                   & Yes        & Yes               \\
% \rowcolor[HTML]{EFEFEF} 
\begin{tabular}[c]{@{}l@{}}Learnable params\end{tabular} & Norm layer             & N/A                   & Prompt       & Norm layer               \\
Model compatibility                                                          & ViTs, CNNs, \etc & ViTs, CNNs, \etc & ViTs & ViTs \& CNNs, \etc \\
\midrule
% \rowcolor[HTML]{EFEFEF} 
Acc. \& Mem. (ViT)                                                                 & 62.7\% (5,166 MB)       & 56.9\% (957 MB)             & 62.4\% (830 MB)   & 66.6\% (825 MB)          \\    
Acc. \& Mem. (ResNet)                                                                 & 46.8\% (5,429 MB)      & 30.9\% (969 MB)             & N/A & 44.9\% (782 MB)         \\
\end{tabular}
}

% \vspace{-8pt}
\label{table:bp_based_free}
\end{table*}

\begin{figure*}[!t]
  % 第一个图 (宽度自适应)
  \begin{minipage}{0.31\textwidth}
    \centering
    \includegraphics[width=\columnwidth]{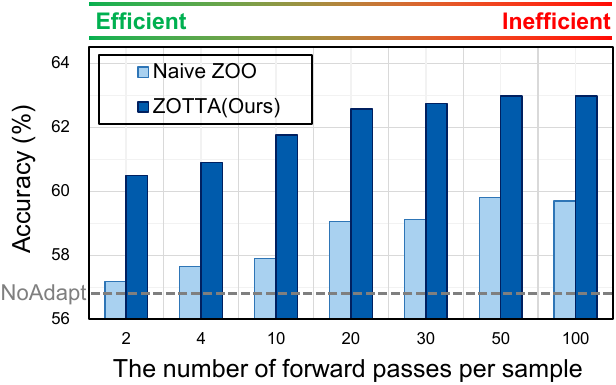}
    % \vspace{-20pt}
    \caption{Comparison of Naive ZOO \vs Ours with varying \#forward passes using ViT-Base on ImageNet-C (Gauss, level 5).}
    \label{fig:naive_zoo}
  \end{minipage}
  \hfill % 弹性间距
  % 第二个图
  \begin{minipage}{0.66\textwidth}
    \centering
    \includegraphics[width=\columnwidth]{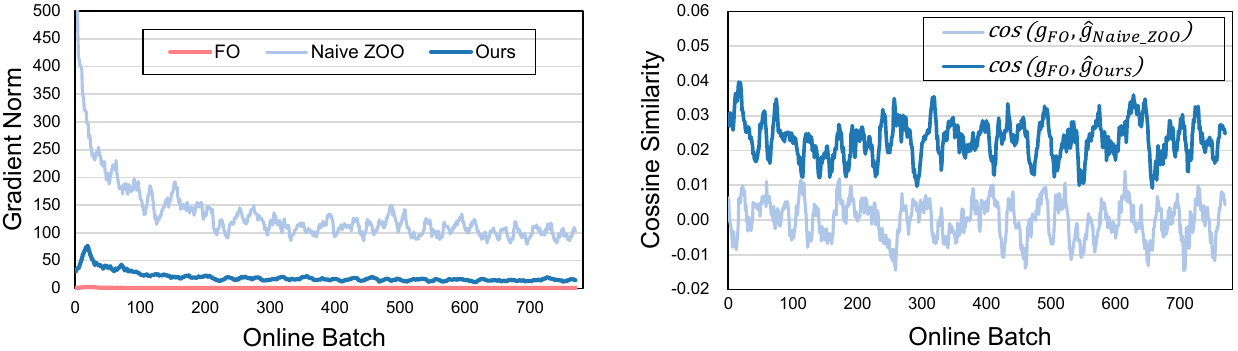}
    % \vspace{-17pt}
    \caption{Analysis of ZOO gradient quality compared to the First-Order (FO) gradient on ImageNet-C, showing (Left) their respective gradient norms and (Right) the cosine similarity between the ZOO gradients (Naive ZOO, Ours) and the FO gradient.}
    \label{fig:grad_analysis}
  \end{minipage}
  % \vspace{-5pt}
\end{figure*}

Zeroth-order optimization (ZOO)~\cite{RGE, liu2020primer, spall2002multivariate, spall1997one} offers a natural path toward addressing the above limitations, providing a theoretically grounded BP-free learning mechanism that estimates gradients through forward-pass perturbations. It requires no backward computation, scales to high-dimensional optimization, and supports arbitrary architectures. However, applying ZOO directly to online unsupervised TTA is inferior in practice, even with extensive forward passes (as shown in Figure~\ref{fig:naive_zoo}). Our diagnostic analysis reveals two key obstacles: \textbf{1) Slow convergence under high-dimensional parameter space}—ZOO’s convergence rate deteriorates as the number of optimized parameters increases, leading to inefficient adaptation; and
\textbf{2) Unstable optimization objective}—without labels, typical entropy-based losses fluctuate heavily under domain shift, leading to noisy gradient estimates and unstable updates.
Hence, the core challenge is not using ZOO itself, but making ZOO efficient and stable for online, unsupervised adaptation.

We propose zeroth-order TTA (ZOTTA), a practical, model-agnostic framework that realizes efficient and stable test-time adaptation purely through forward passes. To tackle the two above challenges, ZOTTA introduces two complementary mechanisms: \textbf{1) Distribution-Robust Layer Selection (DRLS):} Since ZOO’s convergence speed degrades with parameter dimensionality, we propose a layer-wise selection mechanism that identifies and freezes distribution-invariant layers while updating only domain-sensitive ones. This adaptive pruning of the optimization space drastically accelerates convergence without sacrificing adaptation quality. \textbf{2) Spatial Feature Aggregation Alignment (SFAA):} Because ZOO estimates gradients through stochastic perturbations, it is highly sensitive to loss-surface irregularities. SFAA acts as a gradient regularizer, providing a smooth and architecture-agnostic optimization objective by aligning the aggregated feature statistics between source and target domains. This preserves global representational structure, reduces gradient variance, and stabilizes ZOO updates across CNN and ViT architectures. Together, DRLS and SFAA form a synergistic design:
DRLS accelerates ZOO by reducing the optimization space, while SFAA stabilizes ZOO by regularizing the optimization dynamics, jointly enabling efficient and reliable BP-free adaptation.

We validate ZOTTA across diverse benchmarks, including ImageNet-C/R/Sketch/A, and CIFAR100-C, using both CNN and ViT models.
Moreover, we demonstrate its scalability by adapting a multimodal large model on the MathVista benchmark, highlighting its generality and practical deployment value. Our main contributions are as follows:
\begin{itemize}
    \item We introduce ZOTTA, a backpropagation-free TTA framework built upon zeroth-order optimization, enabling efficient and architecture-agnostic adaptation using only forward passes. ZOTTA naturally supports both CNNs and ViTs and is well-suited for resource-constrained non-differentiable deployment scenarios.
    \item We provide diagnostic analysis identifying ZOO's core challenges in TTA—slow convergence under high-dimensional updates and instability from noisy unsupervised objectives—and address them with two components: Distribution-Robust Layer Selection (DRLS) to reduce the optimization dimensionality, and Spatial Feature Aggregation Alignment (SFAA) to provide a stable, low-variance learning signal.
    \item Experiments on ImageNet-C/R/Sketch/A and CIFAR100-C across CNN and ViT backbones show that ZOTTA substantially outperforms prior BP-free methods and even surpasses some BP-based baselines, while drastically reducing memory and computational overhead.

\end{itemize}

\section{Related Work}

\label{sec:related_work}

\textbf{Test-Time Adaptation (TTA)} aims to address the issue of distribution shift during testing through unsupervised or self-supervised methods~\cite{nado2020evaluating, khurana2021sita, boudiaf2022parameter, AME, chencola, cema, wen2023test}. Existing TTA methods can be classified into two categories based on the need for backpropagation (BP). 

\textbf{1) BP-based methods} adapt models during test time by computing gradients via BP. For example, TTT~\cite{TTT} and TTT++~\cite{TTT++} introduce self-supervised tasks, while MEMO~\cite{MEMO} and TENT~\cite{TENT} minimize entropy. SAR~\cite{SAR}, EATA~\cite{EATA} and DeYO~\cite{deyo} focus on selective sample updates. However, BP-based TTA methods incur high computational overhead. While recent methods like EcoTTA~\cite{EcoTTA} and MECTA~\cite{MECTA} reduce memory, they still require far more than BP-free methods and remain reliant on a differentiable framework, limiting real-world applicability.

\textbf{2) BP-free methods} avoid the above issues by eliminating gradient computation. T3A~\cite{T3A} and LAME~\cite{LAME} adjust prototypes or outputs directly, but lack strong learning objectives. To address this, FOA~\cite{foa} introduces a learning-based strategy using evolutionary search over prompts, but is restricted to ViTs and struggles with high-dimensional parameter optimization. In summary, designing a more broadly applicable and efficient learning-based forward-only TTA method is still an open challenge.

\noindent\textbf{Zeroth-Order Optimization (ZOO)} estimates gradients via finite-difference approximations using only forward passes, eliminating the need for backpropagation and offering strong memory efficiency compared to first-order methods like SGD and Adam~\cite{sgd,adam}. Foundational works~\cite{flaxman2004online,ghadimi2013stochastic,duchi2015optimal} have established its theoretical basis, and applications span model reuse~\cite{tsai2020transfer} and model explanation~\cite{dhurandhar2019model}. However, ZOO often suffers from high gradient variance and slow convergence due to its random-direction estimates, limiting its use in deep, high-dimensional models. To improve efficiency, prior works proposed asynchronous parallelism~\cite{lian2016comprehensive}, variance reduction~\cite{liu2018zeroth}, prior-guided sampling~\cite{cheng2021convergence}, and gradient sparsity exploitation~\cite{cai2022zeroth}. Recent methods~\cite{mezo} further extend ZOO to large models.

Although ZOO methods have been largely explored in offline supervised settings. Their applicability to online, unsupervised TTA—where models must adapt to distribution shifts using streaming unlabeled data and limited access per sample—remains underexplored. In this work, we bridge this gap by integrating ZOO into TTA for the first time, leveraging its backpropagation-free nature and memory efficiency to enable model-agnostic online adaptation under distribution shifts.

\section{Methodology}
\label{sec:method}

\paragraph{Problem Statement}
Given a model $f_\theta$ trained on source samples $x \sim P(x)$, test-time adaptation (TTA) aims to adapt it to target samples $x' \sim U(x)$ where $U(x) \neq P(x)$, , to mitigate degradation under distribution shifts. Most TTA methods employ first-order optimization to minimize an unsupervised objective function $\mathcal{L}(\cdot)$, updating the model parameters $\theta$ iteratively as
\begin{eqnarray}
\theta_{t+1} = \theta_t - \eta \cdot \nabla \mathcal{L} (\theta_t),
\label{eqn:update}
\end{eqnarray}
where $\eta$ is the learning rate, and $\nabla \mathcal{L}(\theta_t)$ denotes the gradient of the loss with respect to model parameters $\theta_t$ at time step $t$. However, backpropagation (BP) requires high memory and differentiable computation graphs, making it unsuitable for quantized or resource-constrained cases.
Recent learning-based BP-free methods~\cite{foa} reduce memory cost but typically update only low-dimensional or architecture-specific parameters (e.g., ViT prompts), limiting scalability and generality to different model architectures.

\subsection{Motivation}
\label{sub:zoo}
Zeroth-Order Optimization (ZOO) offers a natural BP-free alternative for TTA, since it estimates gradients using only forward-pass perturbations. Formally, the randomized gradient estimator~\cite{RGE} approximates the gradient via two-sided finite differences:
\begin{eqnarray}
\hat{g} = \frac{1}{k} \sum_{i=1}^{k} \frac{\mathcal{L}(\theta + c u_i ) - \mathcal{L}(\theta - c u_i )}{2c} u_i,
\label{eqn:rge}
\end{eqnarray}
where $k$ represents the number of function queries, with gradient $\hat{g}$ obtained by averaging the $k$ estimations. $u_i \sim \mathcal{N}(0, I)$ denotes a random perturbation sampled from a Gaussian distribution. $c > 0$ is a perturbation size, also referred as the smoothing parameter. 

\textbf{Naive ZOO is insufficient for TTA}
Despite its appealing properties, directly applying ZOO to TTA by updating all normalization parameters (common practice in the TTA field) yields weak adaptation. As in  Figure~\ref{fig:naive_zoo}, ZOO relies on numerous forward passes and achieved limited performance. To understand why, we compare the ZOO-estimated gradient $\hat{g}_{ZOO}$ with a first-order reference gradient $g_{FO}$ computed via BP using the same unsupervised objective.

\textbf{Empirical findings} From Figure~\ref{fig:grad_analysis}, naive ZOO exhibits two persistent issues: 1) Large and volatile gradient magnitude: $||\hat{g}_{ZOO}||$ fluctuates several orders of magnitude more than $||g_{FO}||$, indicating substantial estimator variance. 2) Near-zero gradient alignment: The cosine similarity between $\hat{g}_{ZOO}$ and $g_{FO}$ remains close to zero, suggesting that ZOO updates deviate significantly from the descent direction implied by first-order optimization.

\textbf{Underlying causes} Our analysis reveals two structural factors behind these behaviors: \textbf{1) High estimator variance due to high parameter dimensionality.} ZOO’s finite-difference estimator suffers variance that increases with the dimensionality $d$ of the optimization space. Theory~\cite{RGE} shows that gradient variance scales with $\sqrt{d}$.
In TTA, updating all affine parameters in normalization layers of ViT-base introduces around 38K dimensions, which aligns with the large variance observed in our experiments. \textbf{2) Uniform scalar scaling across all parameters.} The estimator $\hat{g}=\frac{\Delta\mathcal{L}}{2c}u$ uses a single scalar $(\Delta\mathcal{L}/2c)$ to scale the entire $d$-dimensional perturbation vector. This coupling forces all parameters to share one update magnitude, making it difficult for ZOO to differentiate which layers or parameters are truly responsible for correcting domain shift. This leads to noisy and poorly structured updates.

\textbf{Implication for TTA} ZOO theoretically converges at a rate of $O(\sqrt{d}/\sqrt{T})$~\cite{RGE, duchi2015optimal}, meaning that large dimensionality and the small number of optimization steps available in streaming TTA jointly limit its effectiveness. Therefore, to make ZOO effective in online TTA, two key conditions must be met: 1) Reduce the effective optimization dimensionality; 2) Stabilize the unsupervised objective to reduce estimator noise.

\begin{figure*}[!t]
    \centering 
    % \vspace{-10pt}
    \includegraphics[width=\linewidth]{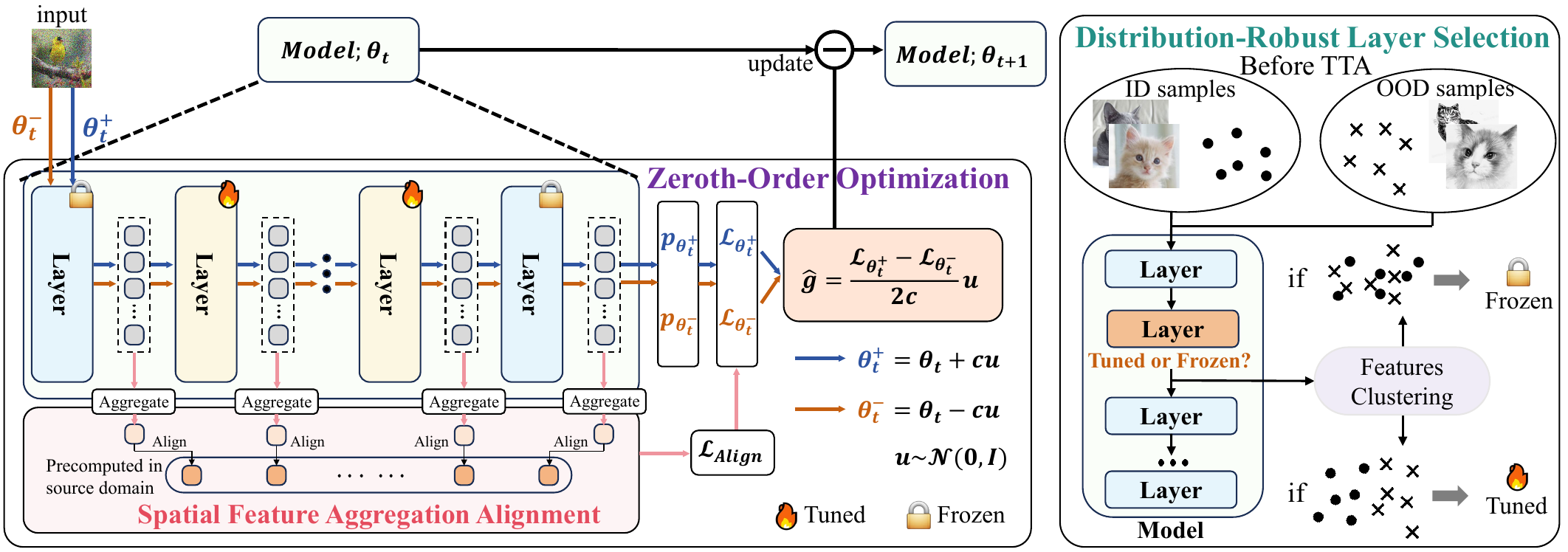}
    % \vspace{-20pt}
    \caption{
    An overview of the proposed ZOTTA framework, which enables BP-free TTA through two main components: 1) \textbf{Distribution-Robust Layer Selection}: Before TTA, we identify distribution-invariant layers and freeze them, updating only distribution-sensitive layers to reduce the optimization dimensionality. 2) \textbf{Zeroth-Order Gradient Estimation via Spatial Feature Aggregation Alignment}: For each TTA step, we inject Gaussian perturbations into the selected parameters, aggregate spatial/token features into global descriptors, and align their statistics with the source domain. This alignment objective provides a more stable zeroth-order gradient obtained from the loss difference of two-sided perturbed forward passes.
} 
% \vspace{-5pt}
    \label{fig:method}
\end{figure*}

\begin{algorithm}[!t]
    \caption{The TTA Pipeline of ZOTTA}
    \label{alg:overall}
    \begin{algorithmic}[1]
        \REQUIRE{Trained model $f(\cdot;\theta)$, target samples set $\mathcal{D}\small{=}\{x_i\}_{i=1}^N$,  
                small sample sets $\mathcal{D}_{id} = \{x_i^{id}\}_{i=1}^K$ and $\mathcal{D}_{ood} = \{x_j^{ood}\}_{j=1}^K$ from ID and OOD domains,
                 purity threshold $\tau$}, hyper-parameter $step$, learning rate $\eta$
        
        \STATE \textbf{Pre-TTA Processing:}
        \STATE \textit{\textbf{// Distribution-Robust Layer Selection:}}
        \STATE Extract layer feature $\mathbf{Z}_i^{id}$ \& $\mathbf{Z}_i^{ood}$ using $D_{id}$ \& $D_{ood}$
        \STATE Compute purity $P_l$ for each layer $l$ via Eqn.~(\ref{eqn:purity}) 
        % \STATE Determine update layer set $\mathcal{U}$
        \STATE Determine update layer set $\mathcal{U} = \{ l \mid P_l \geq \tau \}$
        \STATE \textbf{During TTA:}
        \FOR{$s = 1$ \TO $step$} 
            \STATE \textit{\textbf{// Spatial Feature Aggregation Alignment:}}
            \STATE Calculate the loss $\mathcal{L}$ in Eqn.~(\ref{eqn:tta_loss})
            
            \STATE \textit{\textbf{// Zeroth-Order Gradient Estimation:}}
            \STATE Calculate gradient $\hat{g}$ using $\{x_i\}$ via Eqn.~(\ref{eqn:rge})
            \STATE Update parameters: $\theta \leftarrow \theta - \eta\hat{g}$ 
            \STATE Compute predictions: $\hat{y}_i = f(x_i;\theta)$
        \ENDFOR
    \end{algorithmic}
\end{algorithm}

\subsection{Zeroth-Order TTA Framework}
Based on the above observations, we propose ZOTTA, a BP-free TTA framework that addresses ZOO’s two key limitations. As shown in Figure~\ref{fig:method}, ZOTTA consists of two complementary components:
\textbf{1) Distribution-Robust Layer Selection (DRLS) (Sec.~\ref{sub:drls})}: Select layers to update or freeze based on cross-domain feature separability, reducing trainable parameters and accelerating ZOO convergence.
\textbf{2) Spatial Feature Aggregation Alignment (SFAA) (Sec.~\ref{sub:supervisory_objective})}: Aggregate spatial features into global representations for alignment to reduce domain discrepancies, compensating for the limitations of unstable signals induced by entropy-based TTA objectives.
The overall pseudo-code is summarized in Algorithm~\ref{alg:overall}.

\subsection{Distribution-Robust Layer Selection}
\label{sub:drls}
Although ZOO enables BP-free updates with high memory efficiency, its convergence becomes a bottleneck when adapting high-dimensional models such as CNNs and ViTs—especially in online TTA, where each sample is seen only once. Since ZOO’s convergence rate deteriorates with the number of optimized parameters~\cite{RGE}, reducing the effective optimization dimensionality is crucial for improving adaptation efficiency.

To tackle this, we propose a \textbf{Distribution-Robust Layer Selection} strategy that improves ZOO efficiency by restricting updates to normalization layers that are most sensitive to domain shifts. This strategy is grounded in the motivation that adaptation often seeks to align the feature distribution between source and target domains, \ie, reducing the feature separability between the in-distribution (ID) domain and out-of-distribution (OOD) domain.
In this sense, if the features of ID and OOD data are already aligned before adaptation, further updating becomes less necessary, as such layers contribute little additional information or capacity for adaptation. Conversely, if the ID and OOD features are well separated before adaptation, there exists considerable potential to update these layers, thereby aligning the features within the representation space for adaptation.

To identify such layers, we introduce a lightweight, architecture-agnostic proxy—\emph{clustering purity}—that measures domain separability of layer-wise features and selects only the most domain-sensitive layers for ZOO updates.

Formally, we sample two small sets from both the ID domain and the OOD domain (only 64 samples per domain): 
$D_{id} = \{x_i^{id}\}_{i=1}^{N}$ and $D_{ood} = \{x_i^{ood}\}_{i=1}^{N}$. The OOD samples are obtained at the beginning of the test stream via OOD detection~\cite{berger2021confidence}, while the ID samples can be taken directly from the source data (if available) or extracted from the test stream using the same OOD detection mechanism. Obtaining such small subsets is straightforward in practice and is widely adopted in prior work~\cite{EATA}. Taking ViT as an example, for each sample, we extract features at layer $l$, obtaining feature tensors $\mathbf{Z}^{id}_i \in \mathbb{R}^{M \times d}$ and $\mathbf{Z}^{ood}_i \in \mathbb{R}^{M \times d}$, where $d$ is the feature dimension and $M$ is the number of tokens.
For token $m$, we perform $2$-means clustering on the feature vectors $\{\mathbf{z}_{i,m}^{id}, \mathbf{z}_{i,m}^{ood}\}_{i}$ from all samples within $D_{id}$ and $D_{ood}$. The clustering purity at layer $l$ is defined as:
\begin{equation}
P_{l,m} = \frac{1}{2N} \sum_{q=1}^2 \max_{c \in \{id, ood\}} \left| \mathcal{C}_{q,m} \cap \mathcal{Y}_c \right|,~
P_l = \frac{1}{M} \sum_{m=1}^M P_{l,m},
\label{eqn:purity}
\end{equation}
where $\mathcal{C}_{q,m}$ is the $q$-th cluster and $\mathcal{Y}_c$ is the ground-truth set of features from ID or OOD domain. Here, higher purity reflects stronger domain-specific separation, whereas lower purity indicates distribution invariance due to mixed-domain clustering.
We then select layers for ZOO-based updates using a threshold $\tau$: $\mathcal{U} = \{ l \mid P_l \geq \tau \}$, and freeze all remaining layers. Optionally, shallow layers may also be frozen since they capture generic low-level features (see Sec.~\ref{sec:ablation_select}). This strategy substantially reduces the effective optimization dimensionality, stabilizes ZOO’s gradient estimates, and accelerates convergence—while preserving layers that are already domain-invariant.

\begin{table*}[!t]
\centering
\setlength{\tabcolsep}{3pt} % 将列间距设置为 ..pt
\caption{Comparisons with state-of-the-art methods on ImageNet-C with \textbf{ViT-Base} w.r.t Accuracy(\%).}
% \vspace{-5pt}
% \vspace{-2pt}
\resizebox{\textwidth}{!}{
\begin{tabular}{llcccccccccccccccc}
\multicolumn{1}{l}{}                           & \textbf{}                    & \multicolumn{3}{c}{Noise}                                          & \multicolumn{4}{c}{Blur}                                                           & \multicolumn{4}{c}{Weather}                                                        & \multicolumn{4}{c}{Digital}                                                        &               \\
\multicolumn{1}{l|}{Type}                      & \multicolumn{1}{l|}{Method}  & Gauss.        & Shot          & \multicolumn{1}{l|}{Impul.}        & Defoc.        & Glass         & Motion        & \multicolumn{1}{l|}{Zoom}          & Snow          & Frost         & Fog           & \multicolumn{1}{l|}{Brit.}         & Contr.        & Elastic       & Pixel         & \multicolumn{1}{l|}{JPEG}          & Avg.          \\ \midrule
\multicolumn{1}{l|}{/}                         & \multicolumn{1}{l|}{NoAdapt} & 56.8          & 56.8          & \multicolumn{1}{l|}{57.5}          & 46.9          & 35.6          & 53.1          & \multicolumn{1}{l|}{44.8}          & 62.2          & 62.5          & 65.7          & \multicolumn{1}{l|}{77.7}          & 32.6          & 46.0          & 67.0          & \multicolumn{1}{l|}{67.6}          & 55.5          \\ \midrule
\multicolumn{1}{c|}{\multirow{5}{*}{BP-based}} & \multicolumn{1}{l|}{TENT}    & 60.3          & 61.6          & \multicolumn{1}{l|}{61.8}          & 59.2          & 56.5          & 63.5          & \multicolumn{1}{l|}{59.2}          & 54.3          & 64.5          & 2.3           & \multicolumn{1}{l|}{79.1}          & 67.4          & 61.5          & 72.5          & \multicolumn{1}{l|}{70.6}          & 59.6          \\
\multicolumn{1}{c|}{}                          & \multicolumn{1}{l|}{CoTTA}   & 63.6          & 63.8          & \multicolumn{1}{l|}{64.1}          & 55.5          & 51.1          & 63.6          & \multicolumn{1}{l|}{55.5}          & 70.0          & 69.4          & 71.5          & \multicolumn{1}{l|}{78.5}          & 9.7           & 64.5          & 73.4          & \multicolumn{1}{l|}{71.2}          & 61.7          \\
\multicolumn{1}{c|}{}                          & \multicolumn{1}{l|}{SAR}     & 59.2          & 60.5          & \multicolumn{1}{l|}{60.7}          & 57.5          & 55.6          & 61.8          & \multicolumn{1}{l|}{57.6}          & 65.9          & 63.5          & 69.1          & \multicolumn{1}{l|}{78.7}          & 45.7          & 62.4          & 71.9          & \multicolumn{1}{l|}{70.3}          & 62.7          \\
\multicolumn{1}{c|}{}                          & \multicolumn{1}{l|}{EATA}    & 61.2          & 62.3          & \multicolumn{1}{l|}{62.7}          & 60.0          & 59.2          & 64.7          & \multicolumn{1}{l|}{61.7}          & 69.0          & 66.6          & 71.8          & \multicolumn{1}{l|}{79.7}          & 66.8          & 65.0          & 74.2          & \multicolumn{1}{l|}{72.3}          & 66.5          \\
\multicolumn{1}{c|}{}                          & \multicolumn{1}{l|}{DeYO}    & 62.4          & 64.0          & \multicolumn{1}{l|}{63.9}          & 61.0          & 60.7          & 66.4          & \multicolumn{1}{l|}{62.9}          & 70.9          & 69.6          & 73.7          & \multicolumn{1}{l|}{80.5}          & 67.2          & 69.9          & 75.7          & \multicolumn{1}{l|}{73.7}          & 68.2          \\ \midrule
\multicolumn{1}{c|}{\multirow{4}{*}{BP-free}}  & \multicolumn{1}{l|}{LAME}    & 56.5          & 56.5          & \multicolumn{1}{l|}{57.2}          & 46.4          & 34.7          & 52.7          & \multicolumn{1}{l|}{44.2}          & 58.4          & 61.5          & 63.1          & \multicolumn{1}{l|}{77.4}          & 24.7          & 44.6          & 66.6          & \multicolumn{1}{l|}{67.2}          & 54.1          \\
\multicolumn{1}{c|}{}                          & \multicolumn{1}{l|}{T3A}     & 56.4          & 56.9          & \multicolumn{1}{l|}{57.3}          & 47.9          & 37.8          & 54.3          & \multicolumn{1}{l|}{46.9}          & 63.6          & 60.8          & 68.5          & \multicolumn{1}{l|}{78.1}          & 38.3          & 50.0          & 67.6          & \multicolumn{1}{l|}{69.1}          & 56.9          \\
\multicolumn{1}{c|}{}                          & \multicolumn{1}{l|}{FOA}     & 60.8          & 61.5          & \multicolumn{1}{l|}{62.0}          & 55.7          & 43.5          & 58.5          & \multicolumn{1}{l|}{51.5}          & 66.9          & 64.3          & 70.7          & \multicolumn{1}{l|}{80.5}          & 64.2          & 53.1          & 72.1          & \multicolumn{1}{l|}{70.9}          & 62.4          \\
% \multicolumn{1}{c|}{}                          & \multicolumn{1}{l|}{Vanilla ZO}            & 61.1          & 62.4          & \multicolumn{1}{l|}{62.9}          & 55.4          & 52.5          & 61.3          & \multicolumn{1}{l|}{57.5}          & 67.3          & 64.3          & 67.4          & \multicolumn{1}{c|}{79.9}          & 52.5          & 61.0          & 72.4          & \multicolumn{1}{l|}{71.4}          & 63.3          \\

\multicolumn{1}{c|}{}                          & \multicolumn{1}{l|}{ZOTTA(Ours)}    & \textbf{61.8} & \textbf{63.7} & \multicolumn{1}{l|}{\textbf{63.8}} & \textbf{57.9} & \textbf{57.8} & \textbf{62.8} & \multicolumn{1}{l|}{\textbf{59.7}} & \textbf{70.2} & \textbf{67.8} & \textbf{72.3} & \multicolumn{1}{l|}{\textbf{80.5}} & \textbf{66.0} & \textbf{67.6} & \textbf{74.0} & \multicolumn{1}{l|}{\textbf{73.2}} & \textbf{66.6}
\end{tabular}
}
\label{table:vit_c}
\end{table*}

\subsection{Spatial Feature Aggregation Alignment}
\label{sub:supervisory_objective}
Entropy minimization is widely used in TTA~\cite{TENT,MEMO,TAPR}, but its reliance on model confidence makes it unstable for ZOO. Under distribution shift, prediction logits fluctuate sharply, and ZOO’s finite-difference estimator further amplifies this noise, producing high-variance gradient estimates. Thus, ZOO requires a \emph{smoother, more spatially anchored} objective whose response to perturbations changes more gradually.

\paragraph{Why Spatial Features?}
In both CNNs and Vision Transformers (ViTs), intermediate representations encode rich spatial or token-wise structure: CNNs produce spatial activation maps of size $H\times W\times d$, ViTs produce token embeddings of size $M\times d$. Unlike prediction logits, these spatial/token features evolve smoothly and capture stable, task-relevant statistics (such as activation distribution, magnitude, and global context). Such smoothness makes them ideal for ZOO, whose finite-difference gradient estimator benefits from low-variance objectives. Hence, spatial aggregation provides a natural mechanism to stabilize ZOO.

\paragraph{Unified Spatial Aggregation Across Architectures} To make the objective architecture-agnostic, we unify spatial features from CNNs and ViTs by aggregating them into global descriptors:
\begin{equation}
v_l^{(i)} = h(z_1^{(i)},z_2^{(i)},...,z_m^{(i)}) = \frac{1}{m} \sum_{j=1}^{m} z_j^{(i)},
\end{equation}
where $m$ is the number of spatial maps for CNNs and tokens for ViTs. We implement $h(\cdot)$ as: 1) CNNs--global average pooling over spatial coordinates $(H, W)$; 2) ViTs: averaging over all token embeddings. This spatial aggregation filters local noise, producing stable layer-wise statistics well-suited for ZOO-based adaptation.

\paragraph{Spatial Feature Aggregation Alignment (SFAA)} Based on aggregated features, we align global statistics between the source domain and test-time batches. Before adaptation, for each layer $l$ we compute the source statistics:
\begin{equation}
\mu_l^S = \frac{1}{N} \sum_{i=1}^{N} v_l^{(i)},
\quad
\sigma_l^S = \sqrt{ \frac{1}{N} \sum_{i=1}^{N} \left| v_l^{(i)} - \mu_l^S \right|^2 }.
\end{equation}

During TTA, for the incoming batch $\mathcal{X}_t$, we compute $\mu_l(\mathcal{X}_t)$ and $\sigma_l(\mathcal{X}_t)$ and define the loss:
\begin{flalign}\label{eqn:tta_loss}
     \mathcal{L}&(f(\mathcal{X}_t;\theta)) = \sum_{x \in \mathcal{X}} \sum_{c \in \mathcal{C}_{cls}} -\lambda_1 \hat{y}_c \log \hat{y}_c + \nonumber \\ 
     &\lambda_2 \sum_{l=1}^{L} (\parallel \mu_l(\mathcal{X}_t)-\mu_l^S \parallel^2 + \parallel \sigma_l(\mathcal{X}_t) - \sigma_l^S \parallel^2).
\end{flalign}
This alignment stabilizes ZOO updates by providing a smooth, low-noise objective that anchors intermediate representations to source-domain statistics.

\section{Main Results}
\label{sec:main_results}

\subsection{Experimental Setups}

\textbf{Datasets, Models and Compared Methods}
We conduct experiments on five benchmark datasets: \textbf{1) ImageNet-C}~\cite{imagenet-c} and \textbf{2) CIFAR100-C}~\cite{cifar100}, 15 corruption types across 4 categories, evaluated at severity level 5; \textbf{3) ImageNet-R}~\cite{ImageNet-R}, various artistic renditions of 200 ImageNet classes; \textbf{4) ImageNet-Sketch}~\cite{ImageNet-sketch},  sketch-style images of 1,000 ImageNet classes; \textbf{5) ImageNet-A}~\cite{ImageNet-A}, naturally occurring, real-world adversarial examples.
% 关于实验选用的backbone，我们主要聚焦于ViT-Base，也在ResNet50-GroupNormalization(GN)上进行实验。这两个模型 are trained on the source ImageNet-1K training set。
We primarily focus on ViT-Base~\cite{vit} and also conduct experiments on ResNet50-GN~\cite{resnet, Group_normalization}. Both models are trained on the source ImageNet-1K training set.
The compared SOTA TTA methods include: \textbf{1) BP-based}: TENT~\cite{TENT}, CoTTA~\cite{cotta}, SAR~\cite{SAR}, EATA~\cite{EATA} and DeYO~\cite{deyo}; \textbf{2) BP-free}: LAME~\cite{LAME}, T3A~\cite{T3A} and FOA~\cite{foa}.

\textbf{Implementation Details}
% 附录
We use zeroth-order optimization to estimate gradients. For all backbones, following~\cite{spsa-gc}, we set $k=5$ and $c=0.01$ in Eqn.~(\ref{eqn:rge}). We set the learning rate to 0.01 and clustering purity threshold $\tau$ to 0.6.
For ViT-Base, we set the $\lambda_1$ to 1, and $\lambda_2$ to 0.4. For ResNet50-GN, we set the $\lambda_1$ to 0.1, and $\lambda_2$ to 1.
\textbf{Before TTA}, we acquire 64 samples comprising data from both the source domain and the speckle noise corruption subset of ImageNet-C validation set (simulating data sampled from an online batch). Using these samples, we first calculate statistics for Spatial Feature Aggregation Alignment within the source domain and then perform cluster purity analysis on the model. For ViT-Base, we update parameters in layers 3-5; for ResNet50-GN, we update stage/layer 2.
\textbf{During TTA}, we update only the selected affine transformation parameters of layer normalization for ViT and the group normalization parameters for ResNet50-GN, keeping all other parameters frozen. Notably, $k$ serve as trade-offs between resource usage and efficiency and can be adjusted based on practical requirements.

% 正文
For a fair comparison we set the default $k$ in our method to 5 (i.e., 10 forward passes), while configuring FOA with a population size $K$ of 10 in the main experiments.
For comparisons under varying numbers of forward passes, please refer to Table~\ref{table:runtime} and Figure~\ref{fig:forward_pass}. 
% More details are put in the supplementary material.

\begin{table}[t]
\centering
\caption{Comparisons with state-of-the-arts methods on ImageNet-R,
ImageNet-Sketch and ImageNet-A with ViT-Base w.r.t Acc.(\%).}
\setlength{\tabcolsep}{12pt} % 将列间距设置为 4pt
\resizebox{\columnwidth}{!}{
\begin{tabular}{l|l|ccc|c}
Type                          & Method  & R             & Sketch        & A             & Avg.          \\ \midrule
/                             & NoAdapt & 59.5          & 44.9          & 0.1           & 34.8          \\ \midrule
\multirow{5}{*}{BP-based}     & TENT    & 63.9          & 49.1          & 52.9          & 55.3          \\
                              & CoTTA   & 63.5          & 50.0          & 52.2          & 55.2          \\
                              & SAR     & 63.3          & 48.7          & 52.5          & 54.8          \\
                              & EATA    & 63.3          & 50.9          & 53.4          & 55.9          \\
                              & DeYO    & 66.1          & 52.2          & 54.1          & 57.5          \\ \midrule
\multirow{4}{*}{BP-free} & LAME    & 59.0          & 44.4          & 49.9          & 51.1          \\
                              & T3A     & 58.0          & \textbf{47.6} & 48.9          & 51.5          \\
                              & FOA     & 60.6          & 46.6          & 50.6          & 52.6          \\
                              & Ours    & \textbf{61.5} & 47.3          & \textbf{51.0} & \textbf{53.3}
\end{tabular}
}
\label{table:vit_r/sketch}
\end{table}

\begin{table}[t]
\caption{Resource usage on ViT-Base. \#FP/\#BP indicate forward/backward passes per sample. We report average accuracy on ImageNet-C (level 5). Wall-clock time (s) and memory (MB) are measured on 50,000 images using an RTX4090 GPU. EATA(GC) employs gradient checkpointing to reduce memory. EATA* updates only the norm layers of the last two ViT blocks. $K$ is the population size; $k$ is the number of gradient estimations—both influence \#FP.}
% \vspace{-1pt}
\resizebox{\columnwidth}{!}{
\begin{tabular}{l|l|cc|c|c|c}
Type                      & Method     & \#FP      & \#BP      & ImageNet-C & Runtime(s) & Memory(MB) \\ \midrule
/                         & NoAdapt    & 1         & 0         & 55.5       & 54         & 819        \\ \midrule
\multirow{7}{*}{BP-based} & TENT       & 1         & 1         & 59.6       & 123        & 5165       \\
                          & CoTTA      & 3or35     & 1         & 61.7       & 620        & 16,516      \\
                          & SAR        & {[}1,2{]} & {[}0,2{]} & 62.7       & 243        & 5,166       \\
                          & EATA       & 1         & 1         & 66.5       & 128        & 5,175       \\
                          & EATA(GC)   & 1         & 1         & 66.5       & 226        & 2,701       \\
                          & EATA*      & 1         & 1         & 59.4       & 73         & 1,468       \\
                          & DeYO       & 2         & 1         & 68.2       & 180        & 5,351       \\ \midrule
\multirow{8}{*}{BP-free}  & LAME       & 1         & 0         & 54.1       & 55         & 819        \\
                          & T3A        & 1         & 0         & 56.9       & 126        & 957        \\
                          & Ours(k=1)  & 2         & 0         & 61.7       & 114        & 824        \\
                          & Ours(k=2)  & 4         & 0         & 64.9       & 230        & 824  \\
                          % & FOA(K=10)  & 10        & 0         & 62.4       & 621        & 831        \\
                          % & Ours(k=5)  & 10        & 0         & 66.6       & 559        & 825        \\
                           & \cellcolor[HTML]{EFEFEF}FOA(K=10) & \cellcolor[HTML]{EFEFEF}10 & \cellcolor[HTML]{EFEFEF}0 & \cellcolor[HTML]{EFEFEF}62.4 & \cellcolor[HTML]{EFEFEF}621 & \cellcolor[HTML]{EFEFEF}831 \\
                           & \cellcolor[HTML]{EFEFEF}Ours(k=5) & \cellcolor[HTML]{EFEFEF}10 & \cellcolor[HTML]{EFEFEF}0 & \cellcolor[HTML]{EFEFEF}66.6 & \cellcolor[HTML]{EFEFEF}559 & \cellcolor[HTML]{EFEFEF}825 \\
                          & FOA(K=28)  & 28        & 0         & 66.3       & 1648       & 832        \\
                          & Ours(k=14) & 28        & 0         & 67.6       & 1561       & 828   

\end{tabular}
}

\label{table:runtime}
% \vspace{-5pt}
\end{table}

\begin{table}[t]
% \vspace{-5pt}
% (Group Normalization)
\footnotesize
\centering
\setlength{\tabcolsep}{8pt} % 将列间距设置为 4pt
\caption{Comparisons with state-of-the-art methods on ImageNet-C/R/Sketch/A with \textbf{ResNet50-GN} w.r.t Accuracy(\%).}
% \vspace{-5pt}
\resizebox{\columnwidth}{!}{
\begin{tabular}{l|l|cccc|c}
Type                      & Method  & C             & R             & Sketch        & A             & Avg.          \\ \midrule
/                         & NoAdapt & 31.3          & 40.2          & 28.1          & 13.9          & 28.4          \\ \midrule
\multirow{5}{*}{BP-based} & TENT    & 31.4          & 40.3          & 28.2          & 13.9          & 28.5          \\
                          & CoTTA   & 32.7          & 40.9          & 26.3          & 14.4          & 28.6          \\
                          & SAR     & 46.8          & 51.4          & 36.2          & 20.2          & 38.6          \\
                          & EATA    & 47.1          & 47.7          & 37.5          & 15.9          & 37.0          \\
                          & DeYO    & 34.9          & 41.8          & 28.3          & 14.3          & 29.8          \\ \midrule
\multirow{4}{*}{BP-free}  & LAME    & 30.5          & 40.5          & 28.8          & 13.8          & 28.4          \\
                          & T3A     & 30.9          & 37.0          & \textbf{30.7} & 13.1          & 27.9          \\
                          & FOA     & 24.5          & 39.3          & 26.6          & 12.5          & 25.7          \\
                          & Ours    & \textbf{44.9} & \textbf{43.0} & 30.5          & \textbf{14.1} & \textbf{33.1}
\end{tabular}  
}
% \vspace{-5pt}
\label{table:resnet_c/r/sketch}
\end{table}

\subsection{Comparisons with State-of-the-Art Methods}
\subsubsection{Results using ViT-Base Architecture}
We report the average accuracy of 15 different corruptions on the ImageNet-C dataset (severity level 5) and results on ImageNet-R, -Sketch and -A with ViT-Base in Tables \ref{table:vit_c}, \ref{table:vit_r/sketch} and \ref{table:runtime}.
\textbf{Compared with BP-based methods}, our approach not only surpasses TENT and CoTTA but also outperforms SAR (66.6\% vs. 62.7\%), indicating a BP-free solution can achieve competitive results. While slightly behind the current BP-based SOTA DeYO, our performance can be further improved by increasing the number of forward passes (see Table~\ref{table:runtime}). A major strength of our method is its much lower memory usage (825 MB vs. 5,165 MB for TENT), making it well-suited for resource-limited environments. In contrast, BP-based methods require more memory due to BP and a fully differentiable framework, limiting their applicability on edge devices. Even with optimizations—such as EATA* updating only the last two ViT blocks—performance drops significantly under tighter memory budgets, and memory usage still exceeds that of BP-free methods. \textbf{Compared with BP-free methods}, our approach consistently achieves stronger performance. Notably, it achieves an 8.3\% gain over T3A (66.6\% vs. 56.9\%) and surpasses FOA by 4.2\% (66.6\% vs. 62.4\%) under the same setting of 10 forward passes per sample, without using FOA’s activation shift. This demonstrates our method’s superior adaptation ability with equal computation cost. Moreover, as shown in Table~\ref{table:runtime} and Figure~\ref{fig:forward_pass}(a), our method maintains consistent advantages over FOA across different inference budgets, indicating better efficiency-performance trade-offs. We further validate generalization on ImageNet-R, -Sketch, -A  and CIFAR100-C, where our method again shows strong robustness under diverse distribution shifts. 

\subsubsection{Results using ResNet50-GN Architecture}
\begin{figure*}[!t]
  \centering
  % 第一个图 (宽度自适应)
  \begin{minipage}{0.64\textwidth}
    \centering
    \includegraphics[width=\linewidth]{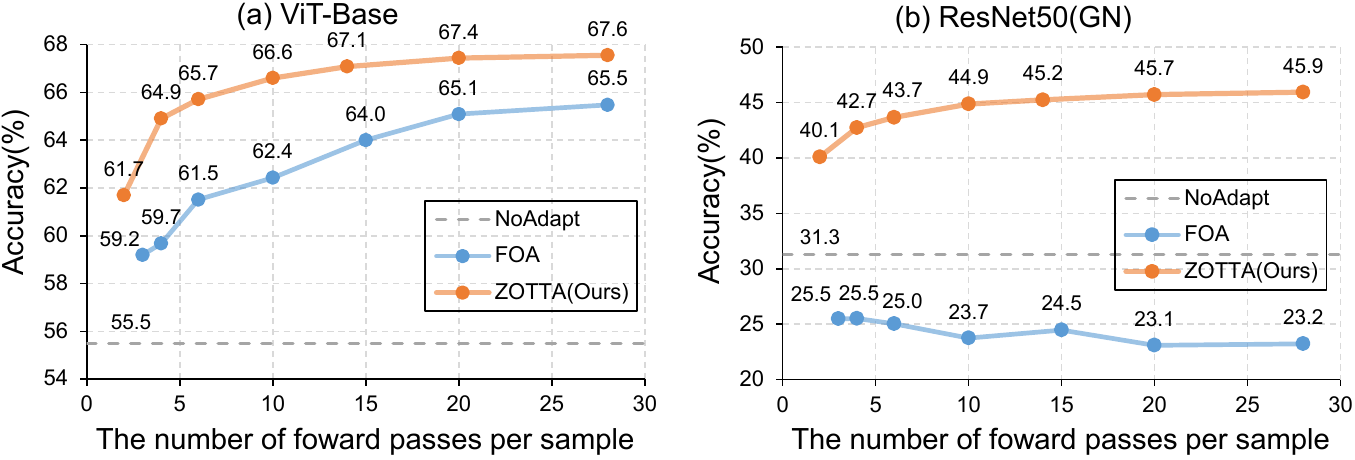} % 第一张图片
    % \vspace{-10pt}
    \caption{Comparison of our ZOTTA vs. FOA under different numbers of forward passes on ImageNet-C (avg acc over level 5 corruptions) using different models.}
    \vspace{-4pt}
    \label{fig:forward_pass}
  \end{minipage}
  \hfill % 弹性间距
  \begin{minipage}{0.32\textwidth}
        \centering
        \includegraphics[width=\textwidth]{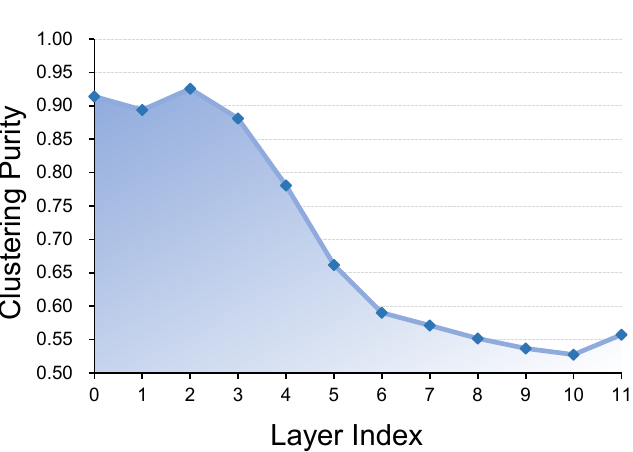} % 第二张图片
        % \vspace{1pt}
        \caption{Per-layer clustering purity of ViT-Base.}
        \label{fig:purity}
  \end{minipage}
\end{figure*}

We report the average accuracy across 15 corruption types on ImageNet-C (severity 5) and results on ImageNet-R, -Sketch and -A using ResNet50-GN (Table~\ref{table:resnet_c/r/sketch}).
\textbf{Compared with BP-based methods}, our method surpasses TENT and CoTTA, and although SAR and EATA remain slightly ahead, we significantly reduce the gap while remaining BP-free—enabling adaptation in non-differentiable or resource-constrained scenarios.
\textbf{Compared with BP-free methods}, our approach shows clear superiority: it outperforms LAME and improves over T3A by a notable 14\% on ImageNet-C (44.9\% vs. 30.9\%).
FOA is hard to generalize beyond ViTs. We attempted to adapt it to ResNet by training a learnable $7\times7$ convolutional prompt added to the input, but it performed worse than no adaptation across multiple datasets—highlighting its poor transferability.
In contrast, our method generalizes well across architectures and, as illustrated in Figure~\ref{fig:forward_pass}(b), consistently outperforms FOA under different numbers of forward passes.

\begin{figure*}[!t]
  \centering
  % 第一个图 (宽度自适应)
  \begin{minipage}{0.32\textwidth}
    \centering
    \includegraphics[width=\columnwidth]{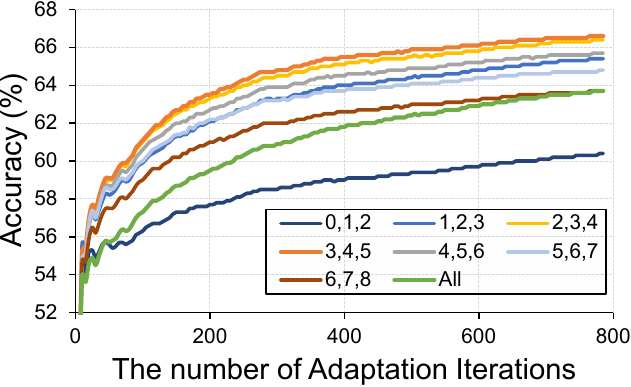}
    % \vspace{-18pt}
    \caption{Convergence rates when updating different layers (numbers denote layer indices selected for update).}
    \label{fig:select_layer}
  \end{minipage}
  \hfill % 弹性间距
  % 第二个图
  \begin{minipage}{0.32\textwidth}
    \centering
    \includegraphics[width=\columnwidth]{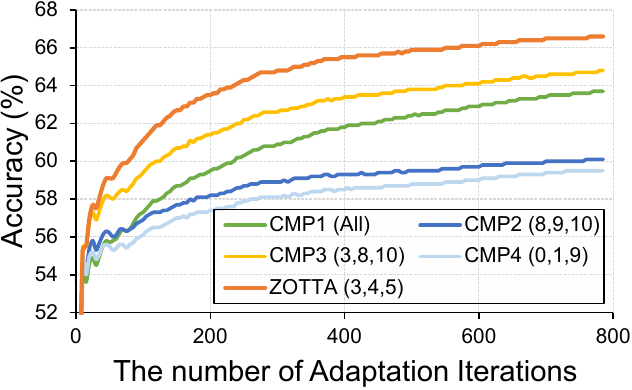}
    % \vspace{-18pt}
    \caption{Convergence rates of different layer selection methods (numbers denote updated layer indices).}
    \label{fig:select_method}
  \end{minipage}
  \hfill
  \begin{minipage}{0.32\textwidth}
    \centering
    \captionof{table}{Ablation of Spatial Feature Aggregation Alignment(SFAA) loss with ViT-Base on ImageNet-C.}
    \begin{tabular}{ccc}
    Entropy & SFAA & Avg. Acc. (\%) \\ \hline
            &      & 55.5           \\
    \checkmark       &      & 60.6           \\
    & \checkmark & 65.5 \\
    \checkmark       & \checkmark    & \textbf{66.6}         
    \end{tabular}
    \label{table:loss}
  \end{minipage}
  \vspace{-5pt}
\end{figure*}

\section{Ablation Studies}
\label{sec:ablation}

\subsection{Effect of Distribution-Robust Layer Selection}
\label{sec:ablation_select}
To analyze the layer-wise distribution robustness, we use 64 pre-stored images from the ImageNet source domain (ID) and 64 images obtained from the first TTA online test batch (OOD) to compute clustering purity at each ViT-Base layer. As shown in Figure~\ref{fig:purity}, purity is high in shallow layers, decreases with depth, and stabilizes around 0.5 in deeper layers—reflecting that shallow layers encode domain-specific features, while deeper layers capture more domain-invariant representations.

To validate the effect of this selection strategy, we conduct an ablation study on ImageNet-C. For each run, we update only three consecutive layers, varying the position of the updated layers across the network. Results in Figure~\ref{fig:select_layer} reveal three observations: 1) Updating layers with high purity (e.g., Layers 3–5, $purity > 0.6$) consistently outperforms full-parameter updates; 2) updating lower-purity layers (e.g., Layers 6–8, $purity < 0.6$) degrades performance; 3) modifying the first layer (Layer 0) leads to noticeable performance drops, likely due to its role in capturing low-level features that should remain unchanged. These findings confirm that clustering purity serves as an effective signal for selecting adaptation-sensitive layers. Moreover, the benefit is not tied to a specific layer: any group of high-purity layers yields better performance than full-parameter updates, validating the flexibility and generality of our strategy.

\noindent \textbf{Comparisons of Different Layer Selection Methods}
% \subsection{Comparison of Layer Selection Methods}
\label{subsec:different_layer}
We compare against four intuitive baselines: 
1) \textbf{CMP1}: Update all parameters;
2) \textbf{CMP2}: Select layers with largest parameter magnitudes;
3) \textbf{CMP3}: Select layers with highest absolute gradient values;
4) \textbf{CMP4}: Identifies layers whose normalization parameters (affine scale/shift) have the greatest influence on outputs. For fair comparison, all methods update exactly three layers during adaptation.
As shown in Figure~\ref{fig:select_method}, our method outperforms all baselines in both accuracy and convergence speed, demonstrating more effective and efficient adaptation than heuristics based on parameter or gradient magnitude. By explicitly evaluating layer-wise domain robustness, it selectively updates distribution-sensitive layers, making it well-suited for unsupervised TTA.

\subsection{Effect of Feature Aggregation Alignment}
% To improve ZOO gradient stability, we combine entropy loss with our proposed model-agnostic Spatial Feature Aggregation Alignment (SFAA). As shown in Table~\ref{table:loss}, entropy alone leads to poor adaptation, while adding SFAA significantly improves performance.
Relying solely on entropy minimization can introduce instability and noisy gradients when combined with ZOO. To improve ZOO's gradient stability, we introduced the Spatial Feature Aggregation Alignment (SFAA) as a model-agnostic objective. As shown in our ablation study (Table ~\ref{table:loss}), using entropy alone leads to poor adaptation (60.6\%), while using SFAA alone achieves a significant performance gain (65.5\%). Combining both objectives yields the best result (66.6\%). This demonstrates that SFAA provides a much more stable learning signal for gradient-free optimization and is critical for improving adaptation robustness.

\subsection{Sensitivity Analysis of Batch Size for SFAA}
\begin{figure}[!t]
    \centering \includegraphics[width=0.8\columnwidth]{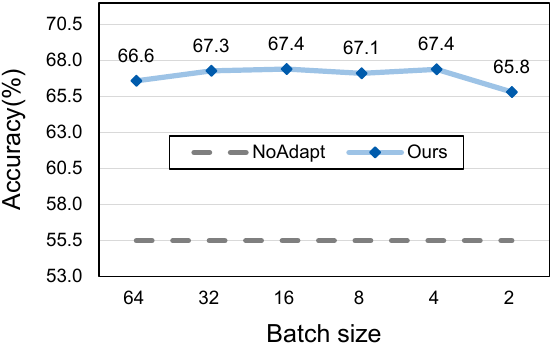}
    \caption{Batch size sensitivity analysis on ImageNet-C using ViT-Base.} 
    \label{fig:batch_size}
\end{figure}

The efficacy of the SFAA loss depends on aligning test-time batch statistics with the source domain. Since the statistical reliability of a batch is inherently sensitive to its size, we conducted an analysis to verify ZOTTA's robustness to small batch sizes. As shown in Figure~\ref{fig:batch_size}, performance remained stable or slightly improved as the batch size was reduced from 64 down to 4, with a significant performance drop only occurring at the extreme minimum of batch size 2. This result confirms that ZOTTA is robust and does not depend on large batches for stable adaptation, as the supervisory signal remains reliable down to a size of 4, although severe statistical bias emerges at the extreme size of 2. To handle the challenge of extremely small batch sizes, such as 1, we employ a special strategy (please refer to Sec.~\ref{sec:wild_setting}).

\subsection{Sensitivity to the Number of Source Samples for SFAA}
\begin{figure}[!t]
    \centering \includegraphics[width=0.8\columnwidth]{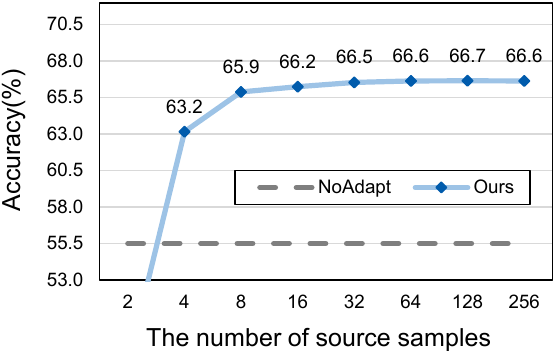}
    \caption{Batch size sensitivity analysis on ImageNet-C using ViT-Base.} 
    \label{fig:source_samples}
\end{figure}

To compute the alignment targets for SFAA, reference statistics from the source domain are required. However, accessing a large source dataset to calculate these global statistics is often impractical in real-world scenarios. We therefore analyze SFAA's sensitivity to the number of source samples used. As shown in Figure~\ref{fig:source_samples}, SFAA is highly data-efficient: using only 8 source samples—a number that is exceptionally practical to obtain—yields a significant accuracy gain (e.g., 65.9\%). Increasing the sample size to 16 (66.2\%), 32 (66.5\%), and 64 (66.6\%) shows diminishing returns, with performance plateauing as sample counts increase further to 128 (66.7\%) and 256 (66.6\%). Considering this saturation point and the balance between practicality and performance, we adopted 64 samples for our default setting, validating SFAA's real-world applicability.

\subsection{Ablation of Weight $\lambda_2$ for SFAA }
\begin{figure}[!t]
    \centering \includegraphics[width=0.8\columnwidth]{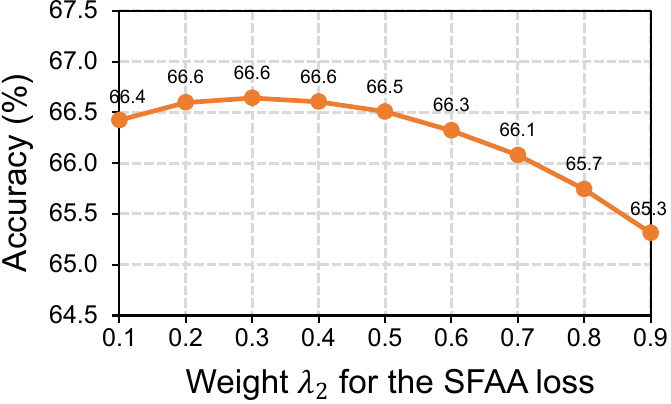}
    \caption{Effect of SFAA loss weight $\lambda_2$ on ImageNet-C using ViT-Base.} 
    \label{fig:ablation_lambda}
\end{figure}
% 我们引入了Spatial Feature Aggregation Alignment (SFAA) objective以在label-free的TTA场景下生成更稳定的梯度。为了探究不同的$\lambda_2$对实验结果的影响，我们固定entropy loss的权重$\lambda_1$为1，在ImageNet-C上使用ViT-Base进行实验，结果如figure~\ref{fig:ablation_lambda}所示。结果表明$\lambda_2$取值在0.1到0.5之间时模型性能稳定，对$\lambda_2$取值并不敏感。但$\lambda_2$取值过大时，模型性能出现下降。
% 这说明当 $\lambda_2$ 过大时，SFAA loss 在优化过程中过度主导了梯度方向，可能干扰且抑制了entropy loss 的作用，导致性能下降。相反，在 $\lambda_2$ 处于 0.1 - 0.5 的合理范围内，SFAA loss 与 entropy loss处在相同数量级,能有效地协同工作.我们在实验中选择 $\lambda_2 = 0.4$ 作为默认值，以平衡两种损失的作用。
We introduce the Spatial Feature Aggregation Alignment (SFAA) objective to generate more stable gradients in label-free TTA scenarios. To investigate the impact of different $\lambda_2$ values in Eqn.~(\ref{eqn:tta_loss}) on experimental outcomes, we fix the weight $\lambda_1$ of the entropy loss at 1.0 and conduct experiments on ImageNet-C using ViT-Base. Results are presented in Figure~\ref{fig:ablation_lambda}. These demonstrate stable model performance when $\lambda_2$ ranges between 0.1 and 0.5, indicating relative insensitivity to $\lambda_2$ within this interval. However, excessive $\lambda_2$ values lead to performance degradation.
The results suggest that an oversized $\lambda_2$ causes the SFAA loss to excessively dominate gradient direction during optimization, potentially interfering with and suppressing the contribution of the entropy loss, thereby impairing performance. Conversely, when $\lambda_2$ resides within the reasonable range of 0.1–0.5, the SFAA loss operates at a comparable scale to the entropy loss, enabling effective collaboration. We select $\lambda_2 = 0.4$ as the default value in our experiments to balance the influence of both loss functions.

\section{Further Experiments and Analysis}
\label{sec:further_experiments}

\subsection{Results on Quantized Models}

Model quantization is commonly employed in real-world deployments to enhance efficiency, but its non-differentiable nature invalidates BP-based TTA methods, making BP-free approaches a more viable solution. Our experiments with 8-bit quantized models (Table~\ref{table:quantized}) show that although quantization degrades the performance of all BP-free methods, our approach achieves 64.8\% accuracy, substantially outperforming the previous best FOA (58.6\%). This demonstrates ZOO maintains robust performance in low-precision scenarios, highlighting its practical value for real-world deployments.

\begin{table*}[!ht]
\caption{Results on quantized ViT-Base models on ImageNet-C with 8-bit floating-point precision w.r.t Accuracy(\%).}
% \vspace{-5pt}
% \vspace{-5pt}
\resizebox{\textwidth}{!}{
\begin{tabular}{lcccccccccccccccc}
                             & \multicolumn{3}{c}{Noise}                                          & \multicolumn{4}{c}{Blur}                                                           & \multicolumn{4}{c}{Weather}                                                        & \multicolumn{4}{c}{Digital}                                                        &               \\
\multicolumn{1}{l|}{Method}  & Gauss.        & Shot          & \multicolumn{1}{c|}{Impul.}        & Defoc.        & Glass         & Motion        & \multicolumn{1}{c|}{Zoom}          & Snow          & Frost         & Fog           & \multicolumn{1}{c|}{Brit.}         & Contr.        & Elastic       & Pixel         & \multicolumn{1}{c|}{JPEG}          & Avg.          \\ \midrule
\multicolumn{1}{l|}{NoAdapt} & 55.3          & 55.1          & \multicolumn{1}{c|}{56.0}          & 45.9          & 34.5          & 51.9          & \multicolumn{1}{c|}{42.1}          & 60.8          & 60.7          & 63.3          & \multicolumn{1}{c|}{77.2}          & 22.3          & 44.2          & 65.9          & \multicolumn{1}{c|}{66.7}          & 53.5          \\
\multicolumn{1}{l|}{T3A}     & 55.6          & 55.7          & \multicolumn{1}{c|}{55.7}          & 45.8          & 34.4          & 51.1          & \multicolumn{1}{c|}{41.2}          & 59.5          & 61.9          & 66.8          & \multicolumn{1}{c|}{76.4}          & 45.5          & 43.4          & 65.6          & \multicolumn{1}{c|}{67.5}          & 55.1          \\
\multicolumn{1}{l|}{FOA}     & 58.3          & 59.0          & \multicolumn{1}{c|}{60.6}          & 52.7          & 36.0          & 54.0          & \multicolumn{1}{c|}{46.5}          & 63.0          & 62.2          & \textbf{71.0} & \multicolumn{1}{c|}{77.6}          & 55.4          & 45.7          & 68.9          & \multicolumn{1}{c|}{68.2}          & 58.6          \\
\multicolumn{1}{l|}{Ours}    & \textbf{60.1} & \textbf{61.8} & \multicolumn{1}{c|}{\textbf{62.0}} & \textbf{56.3} & \textbf{56.4} & \textbf{61.8} & \multicolumn{1}{c|}{\textbf{57.6}} & \textbf{69.2} & \textbf{66.2} & 70.8          & \multicolumn{1}{c|}{\textbf{80.0}} & \textbf{58.8} & \textbf{65.8} & \textbf{73.3} & \multicolumn{1}{c|}{\textbf{72.5}} & \textbf{64.8}
\end{tabular}
}
% \vspace{-2pt}
\label{table:quantized}
\end{table*}

\subsection{Extension to Multimodal Large Model}
To demonstrate ZOTTA's generality and scalability, we extend our evaluation to multimodal large models (MLMs), where massive model size often makes BP-based adaptation impracticala at test time. We apply our BP-free framework to adapt the Qwen2.5-VL-3B~\cite{qwn2.5vl} model on the multimodal reasoning benchmark MathVista~\cite{MathVista}. To simulate a challenging OOD reasoning shift, we focus evaluation on the free-form VQA questions from the three most difficult skill types: ``geometry reasoning'', ``numeric commonsense'', and ``logical reasoning''. During TTA, we employ an entropy minimization objective to update the post-attention normalization layers in the model's decoder. 
As shown in Table~\ref{table:vlm}, ZOTTA's (38.9\%) ability to significantly outperform the NoAdapt (34.7\%) and SOTA BP-free FOA (36.1\%), while approaching the BP-based TENT (39.6\%), confirms its scalability to large-scale multimodal models, highlighting its strong generality and practical value.

\begin{table}[!t]
% \vspace{-5pt}
\footnotesize
\centering
\setlength{\tabcolsep}{9pt} % 将列间距设置为 4pt
\caption{Results on MathVista using the Qwen2.5VL-3B-Instruct model. ``MT-VQA'' denotes math-targeted VQA, and ``G-VQA'' denotes general VQA.}
% \vspace{-5pt}
\resizebox{\columnwidth}{!}{
\begin{tabular}{l|l|cc|c}
% Type                     & Method  & Math-Targeted & General   & Total Acc.    \\ \midrule
Type                     & Method  & MT-VQA & G-VQA   & Total Acc.(\%)    \\ \midrule
/                        & NoAdapt & 51.7              & 22.6          & 34.7          \\ \midrule
BP-based                 & TENT    & 56.7              & 27.4          & 39.6          \\ \midrule
\multirow{2}{*}{BP-free} & FOA     & 53.3              & 23.8          & 36.1          \\
                         & Ours    & \textbf{56.7}     & \textbf{26.2} & \textbf{38.9} \\
\end{tabular}

}
% \vspace{-5pt}
\label{table:vlm}
\end{table}

\subsection{TTA Results on Wild TTA Settings}
\label{sec:wild_setting}
To further validate the robustness of our method in realistic dynamic environments, we evaluate ZOTTA under the ``wild'' TTA settings proposed by SAR~\cite{SAR}. 
These settings simulate three challenging scenarios: \textit{mixed distribution shifts} with randomly shuffled corruptions, \textit{single sample adaptation} (batch size=1), and \textit{imbalanced label shifts} with class-ordered data streams. 
Since our SFAA loss relies on batch statistics, for the single-sample setting, we adopt a lightweight memory queue storing only four historical samples to assist in statistical estimation, following the strategy in FOA~\cite{foa}.
As shown in Table~\ref{table:wild_setting}, ZOTTA consistently outperforms the SOTA BP-free method FOA across all metrics, achieving superior accuracy on mixed shifts (62.8\% vs. 60.8\%), single sample adaptation (67.4\% vs. 64.6\%), and label shifts (62.3\% vs. 60.4\%). 
Moreover, despite being a fully gradient-free approach, ZOTTA achieves performance competitive with, and in some cases surpassing, BP-based methods such as TENT and SAR.
These results demonstrate that ZOTTA maintains exceptional stability and effectiveness even under extreme distribution changes, confirming its suitability for practical edge deployment.

\begin{table}[!t]
% \footnotesize
\centering
\caption{Comparisons with state-of-the-arts methods on ImageNet-C under three type of wild TTA settings with ViT-Base w.r.t Acc.(\%).}
\resizebox{\columnwidth}{!}{
\begin{tabular}{l|l|ccc}
Type                      & Method  & Mixed distribution shift & Batch size=1 & Label shift \\ \midrule
/                         & NoAdapt & 55.5                              & 55.5                  & 55.5                 \\ \midrule
\multirow{4}{*}{BP-based} & TENT    & 60.4                              & 59.3                  & 59.4                 \\
                          & SAR     & 60.7                              & 59.1                  & 63.8                 \\
                          & EATA    & 64.6                              & 66.5                      & 62.5                 \\
                          & DeYO    & 63.9                              & 67.7                      & 64.2                 \\ \midrule
\multirow{2}{*}{BP-free}  & FOA     & 60.8                              & 64.6                  & 60.4                 \\
                          & Ours    & \textbf{62.8}                     & \textbf{67.4}         & \textbf{62.3}       
\end{tabular}
}

\label{table:wild_setting}
\end{table}

\subsection{TTA Results on Continual Setting}
To validate the effectiveness of the proposed method across diverse TTA scenarios, and in accordance with \cite{cotta}, we constructed a more challenging continual adaptation (Continual setting) scenario on ImageNet-C. This scenario simulates the real-world process of continuously evolving environments: data exhibiting distinct corruption types are presented sequentially to the model in complete batches. That is, upon processing all samples of one corruption type, the model immediately commences adaptation to the next type. This setting is designed to evaluate the model's adversarial robustness and long-term adaptation capabilities within dynamically changing conditions. 

Experimental results are detailed in Table~\ref{table:vit_continue} and Table~\ref{table:resnet_continue}.
The results demonstrate that within the continual adaptation setting, the performance of most TTA methods exhibits a decline, underscoring the inherent challenge of this scenario.
\textbf{Comparison with BP-based methods}: Our method significantly outperforms TENT and SAR across both ViT and ResNet architectures. On ViT, its performance (65.0\%) approaches that of DeYO (65.1\%). Notably, EATA achieves optimal performance under this setting; our method, while operating without backpropagation, delivers performance competitive with EATA.
\textbf{Comparison with BP-free methods}: Our method exhibits clear superiority over LAME and T3A. Compared to FOA, our approach not only holds an advantage on ViT (65.0\% vs. 63.9\%) but, more crucially, demonstrates generalizability across diverse model architectures (e.g., ResNet). In contrast, FOA is constrained by its reliance on the ViT architecture and cannot be effectively applied to ResNet.
These results collectively substantiate the proposed method's effectiveness in demanding continual adaptation scenarios, its computational efficiency (being BP-free), and its architectural versatility.

\begin{table*}[!h]
\centering
\caption{Comparisons with state-of-the-arts methods on ImageNet-C under \textbf{continual setting} with ViT-Base w.r.t Acc.(\%).}
\resizebox{\textwidth}{!}{
\begin{tabular}{llcccccccccccccccc}
\textbf{}                                      &                              & \multicolumn{3}{c}{Noise}                   & \multicolumn{4}{c}{Blur}                            & \multicolumn{4}{c}{Weather}                      & \multicolumn{4}{c}{Digital}                          &               \\
\multicolumn{1}{l|}{Type}                      & \multicolumn{1}{l|}{Method}  & Gauss. & Shot & \multicolumn{1}{c|}{Impul.} & Defoc. & Glass & Motion & \multicolumn{1}{c|}{Zoom} & Snow & Frost & Fog  & \multicolumn{1}{c|}{Brit.} & Contr. & Elastic & Pixel & \multicolumn{1}{c|}{JPEG} & Avg.          \\ \midrule
\multicolumn{1}{l|}{/}                         & \multicolumn{1}{l|}{NoAdapt} & 56.8   & 56.8 & \multicolumn{1}{c|}{57.5}   & 46.9   & 35.6  & 53.1   & \multicolumn{1}{c|}{44.8} & 62.2 & 62.5  & 65.7 & \multicolumn{1}{c|}{77.7}  & 32.6   & 46.0    & 67.0  & \multicolumn{1}{c|}{67.6} & 55.5          \\ \midrule
\multicolumn{1}{l|}{\multirow{4}{*}{BP-based}} & \multicolumn{1}{l|}{TENT}    & 57.6   & 59.8 & \multicolumn{1}{c|}{60.9}   & 51.8   & 49.6  & 59.8   & \multicolumn{1}{c|}{53.4} & 64.0 & 62.7  & 68.0 & \multicolumn{1}{c|}{78.6}  & 66.6   & 54.5    & 70.0  & \multicolumn{1}{c|}{69.8} & 61.8          \\
\multicolumn{1}{l|}{}                          & \multicolumn{1}{l|}{SAR}     & 59.2   & 61.2 & \multicolumn{1}{c|}{61.7}   & 54.2   & 55.3  & 58.7   & \multicolumn{1}{c|}{55.7} & 61.0 & 61.9  & 64.6 & \multicolumn{1}{c|}{76.8}  & 58.4   & 58.4    & 68.5  & \multicolumn{1}{c|}{68.8} & 61.6          \\
\multicolumn{1}{l|}{}                          & \multicolumn{1}{l|}{EATA}    & 61.2   & 64.8 & \multicolumn{1}{c|}{65.4}   & 59.1   & 60.3  & 64.7   & \multicolumn{1}{c|}{63.2} & 68.3 & 68.9  & 72.5 & \multicolumn{1}{c|}{79.9}  & 60.8   & 67.5    & 73.3  & \multicolumn{1}{c|}{72.8} & 66.8          \\
\multicolumn{1}{l|}{}                          & \multicolumn{1}{l|}{DeYO}    & 59.8   & 63.5 & \multicolumn{1}{c|}{64.6}   & 56.0   & 57.7  & 62.1   & \multicolumn{1}{c|}{56.4} & 66.5 & 66.2  & 71.7 & \multicolumn{1}{c|}{79.3}  & 65.6   & 63.8    & 72.1  & \multicolumn{1}{c|}{70.9} & 65.1          \\ \midrule
\multicolumn{1}{l|}{\multirow{4}{*}{BP-free}}  & \multicolumn{1}{l|}{LAME}    & 56.5   & 56.6 & \multicolumn{1}{c|}{57.3}   & 46.4   & 34.8  & 52.7   & \multicolumn{1}{c|}{44.2} & 58.4 & 61.6  & 63.1 & \multicolumn{1}{c|}{77.5}  & 24.7   & 44.6    & 66.6  & \multicolumn{1}{c|}{67.2} & 54.1          \\
\multicolumn{1}{l|}{}                          & \multicolumn{1}{l|}{T3A}     & 56.4   & 56.6 & \multicolumn{1}{c|}{56.7}   & 45.3   & 34.5  & 51.8   & \multicolumn{1}{c|}{43.4} & 60.5 & 62.8  & 62.4 & \multicolumn{1}{c|}{77.0}  & 45.7   & 44.4    & 66.7  & \multicolumn{1}{c|}{68.4} & 55.5          \\
\multicolumn{1}{l|}{}                          & \multicolumn{1}{l|}{FOA}     & 60.8   & 63.0 & \multicolumn{1}{c|}{63.9}   & 53.2   & 50.1  & 59.7   & \multicolumn{1}{c|}{56.1} & 65.9 & 70.0  & 70.9 & \multicolumn{1}{c|}{80.8}  & 65.3   & 53.3    & 72.1  & \multicolumn{1}{c|}{72.9} & 63.9          \\
\multicolumn{1}{l|}{}                          & \multicolumn{1}{l|}{Ours}    & 62.2   & 64.7 & \multicolumn{1}{c|}{64.4}   & 55.7   & 57.2  & 61.5   & \multicolumn{1}{c|}{57.9} & 65.8 & 67.8  & 69.3 & \multicolumn{1}{c|}{79.7}  & 64.6   & 62.3    & 71.5  & \multicolumn{1}{c|}{71.1} & \textbf{65.0}
\end{tabular}
}
\label{table:vit_continue}
\end{table*}

\begin{table*}[!h]
\centering
\caption{Comparisons with state-of-the-arts methods on ImageNet-C under \textbf{continual setting} with ResNet50-GN w.r.t Acc.(\%).}
\resizebox{\textwidth}{!}{
\begin{tabular}{llcccccccccccccccc}
\textbf{}                                      &                              & \multicolumn{3}{c}{Noise}                   & \multicolumn{4}{c}{Blur}                            & \multicolumn{4}{c}{Weather}                      & \multicolumn{4}{c}{Digital}                          &               \\
\multicolumn{1}{l|}{Type}                      & \multicolumn{1}{l|}{Method}  & Gauss. & Shot & \multicolumn{1}{c|}{Impul.} & Defoc. & Glass & Motion & \multicolumn{1}{c|}{Zoom} & Snow & Frost & Fog  & \multicolumn{1}{c|}{Brit.} & Contr. & Elastic & Pixel & \multicolumn{1}{c|}{JPEG} & Avg.          \\ \midrule
\multicolumn{1}{l|}{/}                         & \multicolumn{1}{l|}{NoAdapt} & 23.2   & 23.8 & \multicolumn{1}{c|}{23.3}   & 19.1   & 10.9  & 21.0   & \multicolumn{1}{c|}{24.5} & 38.5 & 47.5  & 38.6 & \multicolumn{1}{c|}{68.4}  & 32.3   & 17.9    & 28.2  & \multicolumn{1}{c|}{53.0} & 31.3          \\ \midrule
\multicolumn{1}{l|}{\multirow{4}{*}{BP-based}} & \multicolumn{1}{l|}{TENT}    & 22.2   & 24.4 & \multicolumn{1}{c|}{24.1}   & 20.0   & 13.3  & 23.6   & \multicolumn{1}{c|}{25.4} & 40.5 & 42.7  & 38.1 & \multicolumn{1}{c|}{67.5}  & 38.5   & 17.0    & 36.9  & \multicolumn{1}{c|}{52.0} & 32.4          \\
\multicolumn{1}{l|}{}                          & \multicolumn{1}{l|}{SAR}     & 39.8   & 48.3 & \multicolumn{1}{c|}{47.8}   & 9.1    & 19.6  & 13.9   & \multicolumn{1}{c|}{41.0} & 48.3 & 50.7  & 56.3 & \multicolumn{1}{c|}{70.1}  & 50.6   & 10.1    & 11.0  & \multicolumn{1}{c|}{57.7} & 38.3          \\
\multicolumn{1}{l|}{}                          & \multicolumn{1}{l|}{EATA}    & 40.0   & 43.1 & \multicolumn{1}{c|}{41.0}   & 26.1   & 31.2  & 35.0   & \multicolumn{1}{c|}{42.0} & 45.8 & 47.8  & 53.9 & \multicolumn{1}{c|}{66.6}  & 47.5   & 43.4    & 49.7  & \multicolumn{1}{c|}{55.5} & 44.6          \\
\multicolumn{1}{l|}{}                          & \multicolumn{1}{l|}{DeYO}    & 25.3   & 33.8 & \multicolumn{1}{c|}{36.8}   & 19.5   & 16.9  & 22.6   & \multicolumn{1}{c|}{24.1} & 36.4 & 39.4  & 40.1 & \multicolumn{1}{c|}{62.7}  & 37.7   & 16.8    & 38.0  & \multicolumn{1}{c|}{48.4} & 33.2          \\ \midrule
\multicolumn{1}{l|}{\multirow{4}{*}{BP-free}}  & \multicolumn{1}{l|}{LAME}    & 21.8   & 22.8 & \multicolumn{1}{c|}{21.8}   & 19.6   & 11.1  & 21.1   & \multicolumn{1}{c|}{24.7} & 39.4 & 46.8  & 25.0 & \multicolumn{1}{c|}{68.6}  & 35.9   & 17.5    & 29.0  & \multicolumn{1}{c|}{52.3} & 30.5          \\
\multicolumn{1}{l|}{}                          & \multicolumn{1}{l|}{T3A}     & 20.9   & 22.2 & \multicolumn{1}{c|}{20.0}   & 10.9   & 8.1   & 11.0   & \multicolumn{1}{c|}{19.4} & 33.2 & 39.7  & 28.1 & \multicolumn{1}{c|}{65.9}  & 35.0   & 15.4    & 27.0  & \multicolumn{1}{c|}{50.9} & 27.2          \\
\multicolumn{1}{l|}{}                          & \multicolumn{1}{l|}{FOA}     & 15.8   & 4.0  & \multicolumn{1}{c|}{0.1}    & 0.1    & 0.1   & 0.1    & \multicolumn{1}{c|}{0.1}  & 0.2  & 1.3   & 0.1  & \multicolumn{1}{c|}{4.4}   & 0.1    & 0.2     & 0.4   & \multicolumn{1}{c|}{0.8}  & 1.9           \\
\multicolumn{1}{l|}{}                          & \multicolumn{1}{l|}{Ours}    & 30.5   & 36.3 & \multicolumn{1}{c|}{35.5}   & 22.5   & 25.6  & 36.4   & \multicolumn{1}{c|}{43.7} & 45.4 & 48.7  & 55.3 & \multicolumn{1}{c|}{69.7}  & 45.3   & 43.6    & 52.7  & \multicolumn{1}{c|}{51.8} & \textbf{42.9}
\end{tabular}
}
\label{table:resnet_continue}
\end{table*}

% \subsection{TTA Results on ImageNet-R/Sketch/A}
% To further explore the generalization ability of our method under different types of distribution shifts, we conducte experiments on ImageNet-R and ImageNet-Sketch and ImageNet-A. The results in Table~\ref{table:vit_r/sketch} show that, our method surpasses both LAME and T3A on the average accuracy across the three datasets on ViT-Base. Compared to FOA, our method has a significant advantage (53.3\% vs. 52.6\%), reflecting the robustness of our method when dealing with different out-of-distribution samples. Our method slightly trails BP-based methods, which is expected, as our method does not utilize backpropagation, making the performance gap reasonable.

\subsection{More Results on CIFAR100-C}
To further investigate the generalization capability and universality of the proposed method, we conducted evaluations on the CIFAR100-C dataset in addition to experiments on ImageNet-C and its variants. The results are presented in Table~\ref{table:vit_cifar100_c} and Table~\ref{table:resnet_cifar100_c}.
\textbf{Comparison with BP-based methods}:
Our approach significantly outperforms TENT and SAR across both architectures. For instance, on the ViT architecture, our method (68.9\%) achieves a 4.0\% performance improvement over TENT (64.9\%). Meanwhile, our performance remains comparable to, if not slightly superior to, EATA and DeYO.
\textbf{Comparison with BP-free methods}:
Across both architectures, our method demonstrates substantial advantages over objective-free approaches such as LAME and T3A. Compared with the learning-based BP-free method FOA, our approach exhibits a clear superiority on ViT (68.9\% vs. 63.1\%). More significantly, FOA's design heavily relies on characteristics specific to ViT architecture and fails when applied to ResNet (7.0\%). In contrast, our method delivers robust performance on both ViT and ResNet (60.3\%), further highlighting its exceptional architectural versatility and effectiveness.

\begin{table*}[!ht]
\centering
\caption{Comparisons with state-of-the-arts methods on CIFAR100-C with ViT-Base w.r.t Acc.(\%).}
\resizebox{\textwidth}{!}{
\begin{tabular}{llcccccccccccccccc}
\textbf{}                                      &                              & \multicolumn{3}{c}{Noise}                                          & \multicolumn{4}{c}{Blur}                                                           & \multicolumn{4}{c}{Weather}                                                        & \multicolumn{4}{c}{Digital}                                                        &               \\
\multicolumn{1}{l|}{Type}                      & \multicolumn{1}{l|}{Method}  & Gauss.        & Shot          & \multicolumn{1}{c|}{Impul.}        & Defoc.        & Glass         & Motion        & \multicolumn{1}{c|}{Zoom}          & Snow          & Frost         & Fog           & \multicolumn{1}{c|}{Brit.}         & Contr.        & Elastic       & Pixel         & \multicolumn{1}{c|}{JPEG}          & Avg.          \\ \midrule
\multicolumn{1}{l|}{/}                         & \multicolumn{1}{l|}{NoAdapt} & 44.5          & 47.0          & \multicolumn{1}{c|}{34.0}          & 76.6          & 46.0          & 72.5          & \multicolumn{1}{c|}{80.8}          & 76.6          & 75.2          & 60.4          & \multicolumn{1}{c|}{86.1}          & 53.8          & 66.3          & 53.2          & \multicolumn{1}{c|}{66.6}          & 62.6          \\ \midrule
\multicolumn{1}{l|}{\multirow{4}{*}{BP-based}} & \multicolumn{1}{l|}{TENT}    & 46.0          & 52.2          & \multicolumn{1}{c|}{37.4}          & 76.8          & 47.6          & 73.9          & \multicolumn{1}{c|}{81.8}          & 76.5          & 77.5          & 63.8          & \multicolumn{1}{c|}{85.7}          & 61.2          & 66.8          & 57.8          & \multicolumn{1}{c|}{68.0}          & 64.9          \\
\multicolumn{1}{l|}{}                          & \multicolumn{1}{l|}{SAR}     & 44.6          & 50.8          & \multicolumn{1}{c|}{35.1}          & 75.3          & 43.5          & 72.1          & \multicolumn{1}{c|}{80.4}          & 73.9          & 75.0          & 61.0          & \multicolumn{1}{c|}{84.9}          & 53.6          & 64.9          & 53.8          & \multicolumn{1}{c|}{66.8}          & 62.4          \\
\multicolumn{1}{l|}{}                          & \multicolumn{1}{l|}{EATA}    & 62.8          & 65.9          & \multicolumn{1}{c|}{61.9}          & 80.6          & 65.1          & 78.1          & \multicolumn{1}{c|}{84.1}          & 79.0          & 80.0          & 71.4          & \multicolumn{1}{c|}{87.3}          & 76.5          & 73.4          & 76.9          & \multicolumn{1}{c|}{69.6}          & 74.2          \\
\multicolumn{1}{l|}{}                          & \multicolumn{1}{l|}{DeYO}    & 63.2          & 66.2          & \multicolumn{1}{c|}{64.0}          & 81.7          & 67.0          & 79.7          & \multicolumn{1}{c|}{84.5}          & 80.6          & 81.1          & 74.3          & \multicolumn{1}{c|}{88.0}          & 79.1          & 73.8          & 77.8          & \multicolumn{1}{c|}{71.1}          & 75.5          \\ \midrule
\multicolumn{1}{l|}{\multirow{4}{*}{BP-free}}  & \multicolumn{1}{l|}{LAME}    & 42.2          & 44.8          & \multicolumn{1}{c|}{29.7}          & 75.7          & 43.8          & 71.6          & \multicolumn{1}{c|}{80.1}          & 75.7          & 74.3          & 58.3          & \multicolumn{1}{c|}{85.5}          & 50.9          & 64.9          & 51.4          & \multicolumn{1}{c|}{65.9}          & 61.0          \\
\multicolumn{1}{l|}{}                          & \multicolumn{1}{l|}{T3A}     & 46.4          & 49.9          & \multicolumn{1}{c|}{38.5}          & 77.6          & 53.6          & 73.9          & \multicolumn{1}{c|}{81.3}          & \textbf{77.5} & 76.7          & 63.6          & \multicolumn{1}{c|}{86.6}          & 59.2          & 69.1          & 58.3          & \multicolumn{1}{c|}{\textbf{68.5}} & 65.4          \\
\multicolumn{1}{l|}{}                          & \multicolumn{1}{l|}{FOA}     & 44.6          & 46.9          & \multicolumn{1}{c|}{35.0}          & 76.5          & 49.0          & 72.4          & \multicolumn{1}{c|}{80.6}          & 76.7          & 74.8          & 60.7          & \multicolumn{1}{c|}{85.9}          & 54.1          & 66.6          & 56.9          & \multicolumn{1}{c|}{66.2}          & 63.1          \\
\multicolumn{1}{l|}{}                          & \multicolumn{1}{l|}{Ours}    & \textbf{52.4} & \textbf{56.3} & \multicolumn{1}{c|}{\textbf{51.9}} & \textbf{78.0} & \textbf{59.5} & \textbf{74.6} & \multicolumn{1}{c|}{\textbf{82.3}} & 77.3          & \textbf{77.5} & \textbf{65.7} & \multicolumn{1}{c|}{\textbf{86.7}} & \textbf{63.2} & \textbf{69.5} & \textbf{71.0} & \multicolumn{1}{c|}{67.4}          & \textbf{68.9}
\end{tabular}
}
\label{table:vit_cifar100_c}
\end{table*}

\begin{table*}[!ht]
\centering
\caption{Comparisons with state-of-the-arts methods on CIFAR100-C with ResNet50-GN w.r.t Acc.(\%).}
\resizebox{\textwidth}{!}{
\begin{tabular}{llcccccccccccccccc}
\textbf{}                                      &                              & \multicolumn{3}{c}{Noise}                                          & \multicolumn{4}{c}{Blur}                                                           & \multicolumn{4}{c}{Weather}                                                        & \multicolumn{4}{c}{Digital}                                                        &               \\
\multicolumn{1}{l|}{Type}                      & \multicolumn{1}{l|}{Method}  & Gauss.        & Shot          & \multicolumn{1}{c|}{Impul.}        & Defoc.        & Glass         & Motion        & \multicolumn{1}{c|}{Zoom}          & Snow          & Frost         & Fog           & \multicolumn{1}{c|}{Brit.}         & Contr.        & Elastic       & Pixel         & \multicolumn{1}{c|}{JPEG}          & Avg.          \\ \midrule
\multicolumn{1}{l|}{/}                         & \multicolumn{1}{l|}{NoAdapt} & 17.8          & 19.9          & \multicolumn{1}{c|}{7.2}           & 68.0          & 20.4          & 63.0          & \multicolumn{1}{c|}{73.2}          & 67.8          & 60.3          & 58.3          & \multicolumn{1}{c|}{79.1}          & 67.7          & 49.6          & 32.2          & \multicolumn{1}{c|}{50.4}          & 49.0          \\ \midrule
\multicolumn{1}{l|}{\multirow{4}{*}{BP-based}} & \multicolumn{1}{l|}{TENT}    & 14.3          & 16.6          & \multicolumn{1}{c|}{5.7}           & 69.5          & 19.1          & 64.1          & \multicolumn{1}{c|}{73.7}          & 68.6          & 61.3          & 60.1          & \multicolumn{1}{c|}{79.3}          & 69.4          & 50.7          & 33.1          & \multicolumn{1}{c|}{50.7}          & 49.1          \\
\multicolumn{1}{l|}{}                          & \multicolumn{1}{l|}{SAR}     & 16.2          & 18.2          & \multicolumn{1}{c|}{7.1}           & 69.0          & 19.5          & 63.8          & \multicolumn{1}{c|}{73.7}          & 68.1          & 60.9          & 59.8          & \multicolumn{1}{c|}{79.2}          & 68.9          & 50.6          & 33.0          & \multicolumn{1}{c|}{50.5}          & 49.2          \\
\multicolumn{1}{l|}{}                          & \multicolumn{1}{l|}{EATA}    & 41.1          & 42.2          & \multicolumn{1}{c|}{22.3}          & 72.3          & 41.8          & 69.1          & \multicolumn{1}{c|}{77.1}          & 69.8          & 68.0          & 66.8          & \multicolumn{1}{c|}{79.9}          & 74.7          & 56.8          & 67.9          & \multicolumn{1}{c|}{53.5}          & 60.2          \\
\multicolumn{1}{l|}{}                          & \multicolumn{1}{l|}{DeYO}    & 11.9          & 12.5          & \multicolumn{1}{c|}{3.6}           & 73.0          & 9.1           & 69.5          & \multicolumn{1}{c|}{77.3}          & 70.5          & 66.9          & 65.9          & \multicolumn{1}{c|}{80.3}          & 75.8          & 55.3          & 62.5          & \multicolumn{1}{c|}{52.1}          & 52.4          \\ \midrule
\multicolumn{1}{l|}{\multirow{4}{*}{BP-free}}  & \multicolumn{1}{l|}{LAME}    & 15.8          & 18.1          & \multicolumn{1}{c|}{6.5}           & 67.0          & 18.0          & 60.9          & \multicolumn{1}{c|}{71.7}          & 65.1          & 56.1          & 50.4          & \multicolumn{1}{c|}{78.8}          & 65.6          & 46.6          & 29.4          & \multicolumn{1}{c|}{49.1}          & 46.6          \\
\multicolumn{1}{l|}{}                          & \multicolumn{1}{l|}{T3A}     & 19.1          & 21.7          & \multicolumn{1}{c|}{6.8}           & 66.6          & 22.8          & 61.9          & \multicolumn{1}{c|}{71.5}          & 64.7          & 57.1          & 52.3          & \multicolumn{1}{c|}{77.2}          & 66.3          & 50.7          & 36.5          & \multicolumn{1}{c|}{49.2}          & 48.3          \\
\multicolumn{1}{l|}{}                          & \multicolumn{1}{l|}{FOA}     & 3.8           & 3.8           & \multicolumn{1}{c|}{2.0}           & 54.6          & 1.4           & 3.0           & \multicolumn{1}{c|}{4.1}           & 5.1           & 4.5           & 1.9           & \multicolumn{1}{c|}{7.3}           & 1.1           & 2.5           & 4.1           & \multicolumn{1}{c|}{5.7}           & 7.0           \\
\multicolumn{1}{l|}{}                          & \multicolumn{1}{l|}{Ours}    & \textbf{43.5} & \textbf{44.0} & \multicolumn{1}{c|}{\textbf{22.8}} & \textbf{72.9} & \textbf{42.0} & \textbf{69.3} & \multicolumn{1}{c|}{\textbf{76.0}} & \textbf{69.1} & \textbf{67.2} & \textbf{63.3} & \multicolumn{1}{c|}{\textbf{80.3}} & \textbf{76.2} & \textbf{56.4} & \textbf{68.7} & \multicolumn{1}{c|}{\textbf{53.3}} & \textbf{60.3}
\end{tabular}
}
\label{table:resnet_cifar100_c}
\end{table*}

\subsection{Comparison Analysis with FOA}
Both our proposed method and FOA are BP-free methods with a learning objective, suitable for resource-constrained edge devices. Compared to FOA, our method has two main advantages:
(1) \textbf{Wider applicability}. FOA adapts input prompts which is specifically designed for ViT models. This design limits its generalizability and makes it difficult to apply to CNN architectures. In contrast, our method can be easily applied to various network architectures. (2) \textbf{Fewer dimensional constraints}. FOA uses the covariance matrix adaptation (CMA) method to optimize low-dimensional input prompts, which restricts its ability to directly optimize high-dimensional model parameters across layers and scales. In contrast, the ZOO method we adopt can handle higher-dimensional optimization, offering greater flexibility for adapting to complex models.

\subsection{More Results of Clustering Purity Analysis}

\begin{figure}[t]
    \centering \includegraphics[width=0.8\columnwidth]{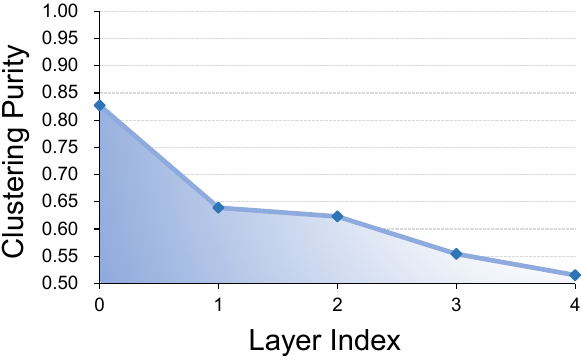}
    \caption{Per-layer clustering purity of ResNet50-GN.} 
    \label{fig:purity_resnet}
\end{figure}

\begin{figure}[t]
    \centering \includegraphics[width=0.8\columnwidth]{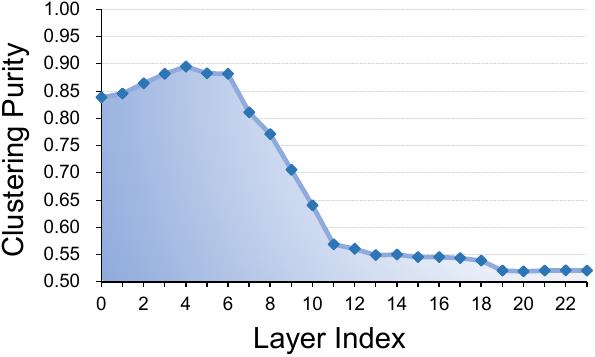}
    \caption{Per-layer clustering purity of ViT-Large.} 
    \label{fig:purity_vit_large}
\end{figure}

Prior to TTA, to analyze the robustness of each model layer to different data distributions, we collect a limited number of samples (e.g., 64 instances) from both the source domain and the speckle noise subset of ImageNet-C (which differs from the test domain) for clustering analysis. These samples comprise both in-distribution (ID) and out-of-distribution (OOD) data. Figure~\ref{fig:purity} presents the clustering analysis results for ViT-Base model. We further extend this analysis to include ResNet50-GN (comprising 5 stages/layers) and ViT-Large (containing 24 layers), with results depicted in Figure~\ref{fig:purity_resnet} and Figure~\ref{fig:purity_vit_large}, respectively. While the curve trends in these figures exhibit minor variations due to architectural differences between models, they consistently demonstrate a declining trend in clustering purity. In later layers, purity falls below 0.6, approaching the theoretical minimum of 0.5. This indicates that these layers map features from different distributions into a shared feature space, effectively extracting distribution-invariant characteristics. Consequently, during TTA, we exclude these layers from updates and instead select alternative layers for adaptation.

\subsection{More Results of Updating Different Layers}
\begin{table}[!t]
\centering
\caption{Results of updating different layers on ImageNet-C with ViT-Base w.r.t Acc.(\%). \underline{Underlined entries} denote results surpassing full-layer update.}
\resizebox{\columnwidth}{!}{
\begin{tabular}{cc|cc|cc|cc}
\begin{tabular}[c]{@{}c@{}}Adapt\\ Layers\end{tabular} & Acc.                & \begin{tabular}[c]{@{}c@{}}Adapt\\ Layers\end{tabular} & Acc.                & \begin{tabular}[c]{@{}c@{}}Adapt\\ Layers\end{tabular} & Acc.                & \begin{tabular}[c]{@{}c@{}}Adapt\\ Layers\end{tabular} & Acc.                \\ \midrule
0                                                      & 57.9                & 0,1                                                    & 59.7                & 0,1,2                                                  & 60.4                & 0,1,2,3                                                & 62.2                \\
1                                                      & 60.0                & 1,2                                                    & 61.6                & 1,2,3                                                  & \underline{65.4}    & 1,2,3,4                                                & \underline{65.6}    \\
2                                                      & \underline{64.5}    & 2,3                                                    & \underline{66.1}    & 2,3,4                                                  & \underline{66.4}    & 2,3,4,5                                                & \underline{\textbf{66.7}} \\
3                                                      & \underline{64.8}    & 3,4                                                    & \underline{\textbf{66.2}} & 3,4,5                                                  & \underline{\textbf{66.6}} & 0,1,2,3,4                                              & 62.9                \\
4                                                      & \underline{\textbf{64.9}} & 4,5                                                    & \underline{65.9}    & 4,5,6                                                  & \underline{65.7}    & 1,2,3,4,5                                              & \underline{66.1}    \\
5                                                      & \underline{63.8}    & 5,6                                                    & \underline{64.7}    & 5,6,7                                                  & \underline{64.8}    & All                                                    & 63.7               
\end{tabular}
}

\label{table:adapt_layers}
\end{table}

% ZOO的收敛速度，与更新的参数量有关。为了加快ZOO的收敛速度，我们提出Distribution-Robust Layer Selection(DRLS)方法，其根据模型每一层对不同分布样本的区分能力（用聚类纯度衡量），冻结能够把不同分布映射到同一特征空间的层，更新剩余的层。我们把聚类纯度阈值$\tau$设为0.6，若某一层的purity小于0.6则会被冻结。在ViT-Base模型中，我们根据阈值冻结了模型的Layer 6到Layer 11。当把剩下的Layer1到Layer5都用来更新时，与更新全部层相比，性能有了明显的提升（66.1% vs. 63.7%），验证了我们方法的有效性。
% 为了探究能否在阈值划定的范围内继续减少更新的层数以进一步提高ZOO收敛速度，我们在ImageNet-C上进行了实验。table~\ref{table:adapt_layers}展示了更新不同数量的层的实验结果。主要有下面3点observation。（1）无论仅更新一层或是更新多层，都能够优于更新全部层的结果。这说明在purity阈值筛选选出来的层中选少量或多数层更新均可以，验证了选层方法的有效性。（2）选不同数量的层更新的最优结果有差距。更新3层或4层时结果最优，能达到66.6%，而更新1层时最好结果只有64.9%。说明更新多少层需要权衡，更新太多会导致收敛速度慢，而更新太少则使得模型adapt能力有限。（3）选择浅层尤其是Layer 0更新时效果较差，原因可能时最浅的层往往提取的是低维的通用的信息（如物体的边缘、轮廓），这些层通用性强，但更新较为困难，应该使其保持冻结。

The convergence speed of Zeroth-Order Optimization (ZOO) is influenced by the volume of parameters updated. To accelerate ZOO convergence, we propose the Distribution-Robust Layer Selection (DRLS) method. This approach freezes layers capable of mapping different distributions to a homogeneous feature space while updating the remaining layers, based on each layer’s distribution-discriminative capacity—quantified by its clustering purity.
We set the clustering purity threshold $\tau$ to 0.6. layers with purity below this threshold are frozen. For the ViT-Base model, Layers 6 to 11 were frozen accordingly. Updating only Layers 1 to 5 yielded a marked performance improvement (66.1\% vs. 63.7\%) compared to full-model updates, validating the efficacy of our method.

To explore whether further reducing updated layers within the threshold-defined range could enhance ZOO convergence, experiments were conducted on ImageNet-C. Table~\ref{table:adapt_layers} presents results for updating varying numbers of layers, revealing three key observations:
(1) Updating either single or multiple layers within the purity-filtered range consistently outperformed full-layer updates, confirming the validity of our layer selection strategy.
(2) Optimal results varied with the number of updated layers: Performance peaked at 66.6\% when updating 3 or 4 layers, whereas updating only 1 layer yielded a maximum of 64.9\%. This indicates a trade-off—excessive layers slow convergence, while insufficient layers limit adaptation capacity.
(3) Shallow layers (notably Layer 0) empirically exhibited poorer performance when updated. This is likely attributable to their role in extracting low-dimensional, general features (e.g., edges, contours), which exhibit high transferability but are less amenable to adaptation and should remain frozen.

\subsection{Visual Analysis of Cross-Distribution Feature after Model Adaptation}

\begin{figure*}[!t]
    \centering \includegraphics[width=\textwidth]{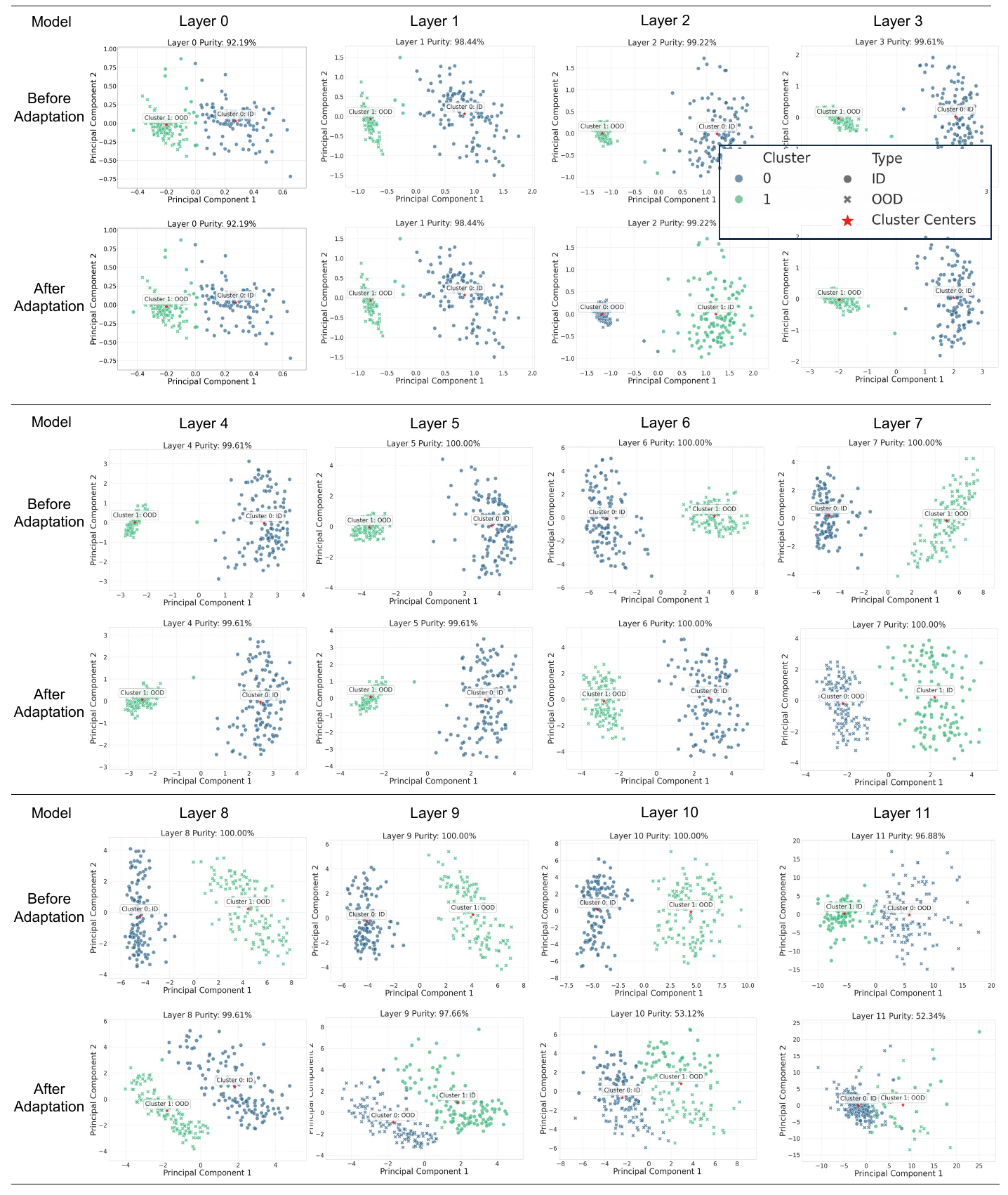}
    \caption{Per-layer clustering results of CLS token for model before and after adaptation on ImageNet-C gaussian noise using ViT-Base.} 
    \label{fig:visual}
\end{figure*}

% 为了验证使用我们proposed方法 xxx（补充），我们对adapt前后的模型的每一层输出进行可视化分实验，探究我们的方法为模型带来哪些变化。具体来说，我们从源域和ImageNet-C的gaussian noise分别取128个ID和OOD样本，将这些样本混合后，输入到标准的ViT-Base模型及其经我们方法适应（adapt）后的版本中。对于模型每一层输出的CLS token特征，我们应用K-Means聚类算法（K=2）进行分组，旨在观察模型在不同深度对ID/OOD样本的区分能力变化}展示了模型adapt前后各层CLS token的聚类的可视化结果(使用PCA方法降维)。
% 在未经适应的原始模型中，从浅层到深层，几乎所有层的CLS token特征都能将ID样本和OOD样本清晰地区分开来。这表明原始模型在各个层次都倾向于提取能够区分不同数据分布的特征信息。而模型用我们proposed方法adapt以后，主要有以下两点observation。
% （1）浅层保留区分性 (Layer 0-~7)： 在模型的较浅层（如前7层），ID和OOD样本的特征聚类结果与适应前类似，依然保持较好的分离状态。这表明浅层特征主要捕获低级、通用的图像结构信息，这些信息受特定领域分布偏移的影响相对较小
% （2）深层开始特征趋同 (Layer 8-~11)： 从第8层左右开始，一个显著的趋势出现：ID样本和OOD样本的特征簇开始逐渐靠近。这暗示了模型在中间层到深层的学习过程中，正在被引导去忽略那些导致分布偏移（如ImageNet-C中的高斯噪声）的特定因素。 值得注意的是，在模型的最终输出层（第11层），ID和OOD样本的CLS token特征几乎完全重叠在一起，形成单一且密集的簇，难以通过简单的聚类区分开来。

% 实验结果说明， 深层特征的重叠现象并非模型性能退化，相反，它揭示了模型成功学习到了分布不变（Distribution-Invariant）的特征表示。能够把来自不同分布的特征映射到一个一致的特征空间。即捕捉到跨越不同分布的、与核心语义任务（如图像分类）相关的本质信息，而过滤掉了导致域间差异（如特定噪声模式）的干扰因素。
% 因此，本可视化实验不仅验证了我们的方法能有效驱动模型学习分布不变特征，还清晰地阐明了这一学习过程在模型层级结构中的具体演变路径（从浅层保留区分性到深层实现特征一致性），为方法的有效性提供了深层次的洞察。
To investigate the transformations induced by our approach, we conducted a visualization experiment analyzing layer-wise outputs of the model before and after adaptation. Specifically, we sampled 128 in-distribution (ID) samples from the source domain and 128 out-of-distribution (OOD) samples from ImageNet-C’s Gaussian noise corruption, combining them as input to both the standard ViT-Base model and its adapted counterpart. For the CLS token features extracted from each layer, we applied K-Means clustering (K=2) to examine the model’s evolving capacity to distinguish ID/OOD samples across depths. The visualization of clustering results (using PCA for dimensionality reduction) is presented in figure~\ref{fig:visual} for both pre- and post-adaptation models.

In the unadapted baseline model, CLS token features across nearly all layers—from shallow to deep—exhibited clear separation between ID and OOD samples. This indicates that the original model inherently extracts features capable of discriminating between distinct data distributions at every hierarchical level. After adaptation with our method, two primary observations emerge:
(1)\textbf{ Preservation of Discriminability in Shallow Layers} (Layers 0–5):
In early layers, the clustering results for ID and OOD features remain comparable to those before adaptation, maintaining well-separated groupings. This suggests that shallow-layer features predominantly capture low-level, generalizable image structural information.
(2) \textbf{Progressive Feature Convergence in Deep Layers} (Layers 6–11):
From approximately Layer 6 onward, a pronounced trend emerges: the feature clusters of ID and OOD samples gradually converge. This implies that during mid-to-deep layer processing, the model is guided to disregard factors inducing distribution shifts. Notably, at the final output layer (Layer 11), the CLS token features of ID and OOD samples become nearly indistinguishable after adaptation, forming a single dense cluster that cannot be separated through simple clustering.

These experimental results demonstrate that the feature overlap in deep layers does not indicate performance degradation. Rather, it reveals the model’s successful acquisition of distribution-invariant feature representations. Our approach effectively maps features from divergent distributions into a unified embedding space—preserving semantic information essential to the core task (e.g., image classification) while filtering out domain-specific perturbations.
Consequently, this visualization study not only verifies our method’s efficacy in driving models to learn distribution-invariant features but also elucidates the evolutionary trajectory of this learning process across the model hierarchy: from discriminability preservation in shallow layers to feature alignment in deep layers. This provides profound insights into the methodological validity.

% 值得注意的是，这里仅展示了CLS token的聚类结果。所有token的聚类结果可参考figure 3。由于CLS token更具判别力，从Layer 8开始才出现聚类纯度下降。而对于所有token而言，平均聚类纯度在Layer6已经下降到0.6以下，说明已具备提取distribution-invariant特征的能力。因此在选择更新的层时，我们主要根据全部token的聚类结果而不是CLS token。
% It is noteworthy that only the clustering results of the CLS token are presented here. Refer to Figure~\ref{fig:purity} for clustering outcomes of all tokens. Given the superior discriminative power of the CLS token, a decline in clustering purity does not manifest until Layer 8. However, for all tokens collectively, the average clustering purity falls below 0.6 as early as Layer 6, indicating the emergence of capabilities for extracting distribution-invariant features. Consequently, our selection of layers for updating is primarily guided by the clustering performance of all tokens rather than that of the CLS token.

\section{Conclusion}
\label{sec:conclusion}
We have proposed ZOTTA, a practical backpropagation-free test-time adaptation framework that enables online, unlabeled adaptation using forward passes only. In making ZOO effective for TTA, we have identified two key challenges—slow convergence under high-dimensional updates and instability from noisy unsupervised objectives—and have addressed them with two complementary components. Distribution-Robust Layer Selection (DRLS) has reduced the effective optimization dimensionality, while Spatial Feature Aggregation Alignment (SFAA) has provided a smooth, low-variance supervisory signal tailored to ZOO, jointly ensuring efficient and stable adaptation across architectures. ZOTTA has achieved superior performance across ImageNet-C/R/Sketch/A and CIFAR100-C on both ViTs and CNNs, surpassing prior BP-free methods and rivaling BP-based baselines with much lower cost, making it a scalable and deployment-ready solution for real-world TTA.

\bibliographystyle{IEEEtran}
\bibliography{main}

\vfill
\end{document}